\documentclass{article}

\IfFileExists{build-mode.tex}{\def\arxivmode{}\def\arxivaccepted{}
}{}
\PassOptionsToPackage{numbers, compress}{natbib}

\newif\ifpublicversion
\ifdefined\arxivmode
  \publicversiontrue
  \ifdefined\arxivpublished
    \usepackage[eandd,final]{venue/neurips_2026}
  \else
    \usepackage[preprint]{venue/neurips_2026}
  \fi
\else
  \ifdefined\cameraready
    \publicversiontrue
    \usepackage[eandd,final]{venue/neurips_2026}
  \else
    \usepackage[eandd]{venue/neurips_2026}
  \fi
\fi

\usepackage[utf8]{inputenc}
\usepackage[T1]{fontenc}
\usepackage{hyperref}
\usepackage{url}
\usepackage{booktabs}
\usepackage{amsfonts}
\usepackage{nicefrac}
\usepackage{microtype}
\usepackage{xcolor}
\usepackage{xcolor}
\usepackage{graphicx}
\usepackage[most]{tcolorbox}
\usepackage{caption}
\usepackage{subcaption}
\usepackage{wrapfig}
\usepackage{float}
\usepackage{colortbl}
\usepackage{multirow}
\usepackage{threeparttable}
\usepackage{pifont}
\usepackage{enumitem}
\newtcolorbox{promptbox}[1]{
  enhanced,
  breakable,
  colback=white,
  colframe=gray!75!black,
  boxrule=1.2pt,
  sharp corners,
  left=8pt,
  right=8pt,
  top=8pt,
  bottom=8pt,
  before skip=10pt,
  after skip=10pt,
  fonttitle=\bfseries,
  title={#1}
}
\newenvironment{promptboxcap}[2]
  {\def\promptboxcaption{#2}\begin{promptbox}{#1}}
  {\end{promptbox}\captionof{figure}{\promptboxcaption}}

\IfFileExists{build-mode.tex}{%
  \newcommand{\sharedroot}{.}%
}{%
  \newcommand{\sharedroot}{../..}%
}
\IfFileExists{build-mode.tex}{%
  \graphicspath{{\sharedroot/figures/}}%
}{%
  \graphicspath{{\sharedroot/build/figures/}{\sharedroot/figures/}}%
}

\newcommand{\name}{{MDMR-Bench}}
\newcommand{\dataseturl}{https://huggingface.co/datasets/MDMR-Bench/MDMR-Bench}
\ifpublicversion
  \newcommand{\codeurl}{https://github.com/LHRYANG/MDMR-Bench}
\else
  \newcommand{\codeurl}{https://anonymous.4open.science/r/Neurips2026_ED_paper3412/}
\fi

\title{\name{}: A Multi-Dimensional Benchmark for Multi-Reference Image Generation}
\ifpublicversion
\author{
\textbf{Bowen Zheng$^{1\ast}$ \quad
Jintao Lin$^{2\ast}$ \quad
Changheng Yi$^{1}$ \quad
Haoran Yang$^{3\dagger}$} \\
\textbf{Binxiao Huang$^{3}$ \quad
Chenyang Lei$^{3}$ \quad
Rui Liu$^{3}$ \quad
Xihui Liu$^{2\dagger}$} \\
\\
$^{1}$Nanjing University \quad
$^{2}$The University of Hong Kong \quad
$^{3}$Huawei Research
}

\else
  \author{Anonymous authors}
  \hypersetup{pdfauthor={}}
\fi

\ifdefined\arxivaccepted
  \makeatletter
  \renewcommand{\@noticestring}{Accepted at NeurIPS 2026, Evaluations \& Datasets Track.}
  \makeatother
\fi

\begin{document}
\maketitle
\ifpublicversion
  \begingroup
  \renewcommand{\thefootnote}{\ensuremath{\ast}}
  \footnotetext{Equal contribution.}
  \renewcommand{\thefootnote}{\ensuremath{\dagger}}
  \footnotetext{Corresponding authors: \texttt{eurekayang96@outlook.com}, \texttt{xihuiliu@eee.hku.hk}.}
  \endgroup
\fi
\begin{figure}[h]
\vspace{-6ex}
    \centering
    \includegraphics[width=\linewidth]{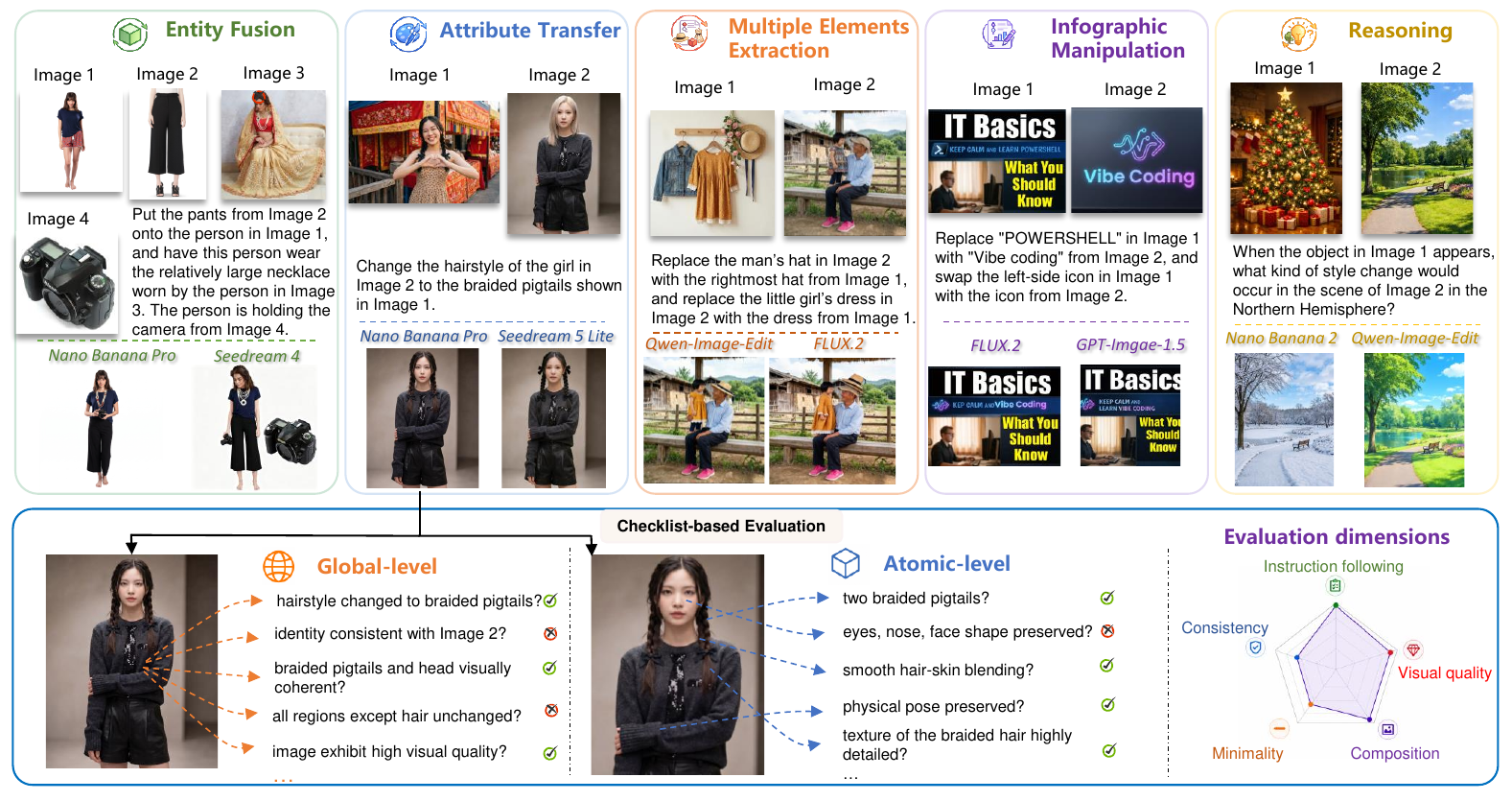}
    \caption{The overview of \name{}. It covers diverse challenge dimensions for multi-reference image generation and introduces a checklist-based evaluation framework across five evaluation aspects with global-level and atomic-level queries.}
    \label{fig:teaser}
\end{figure}

\begin{abstract}
Image generation models have rapidly advanced, with many models now supporting generation conditioned on multiple reference images and textual instructions. Such models enable a wide range of applications, including entity fusion and attribute transfer across images. However, existing benchmarks for multi-reference image generation rely on limited evaluation dimensions and fail to capture the diverse challenges of the task.
To address this limitation, we propose \textbf{MDMR-Bench}, a \textbf{M}ulti-\textbf{D}imensional benchmark for \textbf{M}ulti-\textbf{R}eference image generation. MDMR-Bench is organized around representative capability settings and consists of multiple evaluation subsets spanning diverse dimensions of multi-reference image generation, including (1) common entity fusion and attribute transfer, (2) multi-element extraction from a single image, (3) infographic manipulation, and (4) visual reasoning-intensive generation. This design enables systematic analysis of model capabilities and failure modes across different challenge dimensions. Furthermore, to achieve more reliable and interpretable evaluation, we introduce a checklist-based evaluation protocol that decomposes generation quality into fine-grained evaluation aspects. Instead of relying solely on holistic scoring, our framework uses structured checklists composed of both global-level and atomic-level queries, enabling more diagnostic assessment of model performance. By benchmarking both open-source and proprietary models, we demonstrate that MDMR-Bench provides fine-grained insights into model performance across different dimensions.
The dataset is publicly available at Hugging Face\footnote{\href{\dataseturl}{\name{} dataset on Hugging Face}}.
The code is released at GitHub\footnote{\href{\codeurl}{\name{} code repository on GitHub}}.
\end{abstract}

\section{Introduction}
\label{sec:introduction}
Text-to-image~\cite{blackflux2024,wu2025qwenimagetechnicalreport,chen2025blip3o,esser2024sd3,team2025zimage} and instruction-based single-image editing models~\cite{wang2025seededit30fasthighquality,liu2025step1x-edit,blackforestlabs2025kontext,LongCat-Image,firered2026rededit} have achieved remarkable progress. They support high-quality image generation from text and fine-grained, controllable edits on a single image. However, text alone is often insufficient to precisely specify complex visual attributes, such as the exact appearance of clothing, or abstract properties like hairstyle, pose, and artistic style. In practice, users naturally rely on multiple reference images to convey such details, leading to a growing demand for multi-reference image generation. In response, recent models, such as Qwen-Image-Edit ~\cite{wu2025qwenimagetechnicalreport} and FLUX.2~\cite{blackforestlabs2025flux2}, as well as proprietary models like GPT-Image-1.5~\cite{openai2025gptimage15}, Seedream~\cite{seedream2025seedream,seedream2026seedream5lite}, and Nano-Banana Pro~\cite{nanobananapro2025}, support multi-image inputs and demonstrate promising capabilities in integrating information from them.

While benchmarks for text-to-image generation (e.g., GenEval~\cite{ghosh2023geneval,kamath2025geneval2}, DPGbench~\cite{hu2024dpgbench}) and instruction-based image editing (e.g.GEdit~\cite{liu2025step1x-edit, jiang2026geditv2}, ImgEdit~\cite{ye2025imgedit}) have become increasingly rich and comprehensive, benchmarks for multi-reference image generation remain scarce. Representative efforts have begun to explore this setting, with several benchmarks recently introduced. However, as shown in Table~\ref{tab:intro_comparision}, existing benchmarks remain limited in several aspects. First, some existing benchmarks are limited in scale, often containing only a few hundred cases~\cite{xia2025dreamomni2, wu2025omnigen2}, and provide relatively narrow task coverage. Many primarily focus on simple fusion or transfer scenarios, while some are restricted to specific applications, such as virtual try-on in VTEdit-Bench~\cite{liang2026vtedit}. Second, reference images are often overly simple and contain many visually unnatural samples, typically dominated by salient foreground objects, and thus do not reflect the complexity of real-world scenes~\cite{wu2025omnigen2, wei2025mico, xia2025dreamomni2, oshima2025multibanana}. Most importantly, existing benchmarks lack a multi-dimensional evaluation settings, typically organizing tasks along only a single aspect, such as the number of references or coarse task categories. While some benchmarks vary reference properties (e.g., photo fused with anime)~\cite{oshima2025multibanana}, such variations mainly reflect superficial input styles rather than the diverse capabilities required in real-world multi-reference image generation.

\begin{wrapfigure}{r}{0.42\textwidth}
    \vspace{-0.8em}
    \centering
    \includegraphics[width=0.40\textwidth]{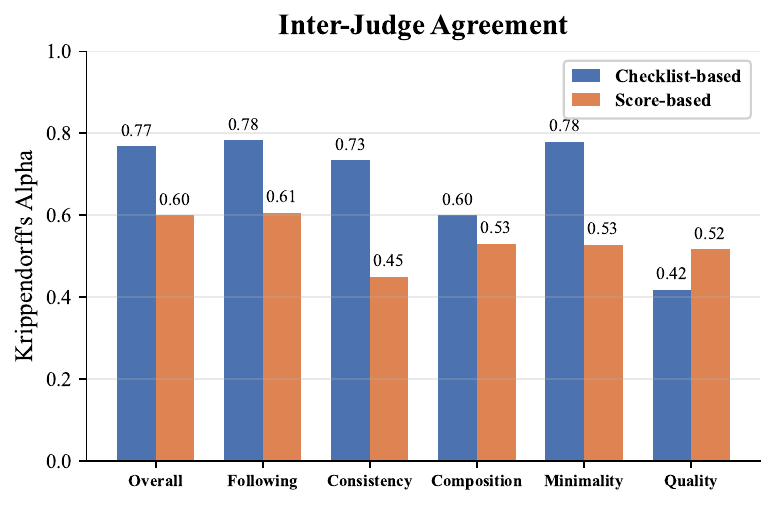}
    \caption{Inter-judge agreement measured by Krippendorff's alpha. Settings are provided in the Appx.~\ref{appx:inter_judge}}
    \label{fig:inter_anno_alignment}
    \vspace{-1.0em}
\end{wrapfigure}

To address these limitations, we propose \textbf{MDMR-Bench}, a \textbf{M}ulti-\textbf{D}imensional benchmark for \textbf{M}ulti-\textbf{R}eference image generation, containing 2,390 cases collected from visually realistic sources, including both real-world images and high-quality synthetic images, as illustrated in Figure~\ref{fig:teaser}. Its main contribution is capability diversity: dedicated coverage of under-evaluated task families, global- and atomic-level checklist evaluation across five diagnostic dimensions, and detailed analysis of model capabilities and failure modes. Although MDMR-Bench is smaller than MultiBanana and VTEdit-Bench by raw sample count, it provides broader dedicated task coverage than MultiBanana, as shown in Table~\ref{tab:intro_comparision}. In addition to varying the number of reference images, we include scenarios requiring multi-element extraction from a single reference image, infographic manipulation involving visual elements such as artistic text and logos, and reasoning-intensive generation that requires implicit relationship understanding across multiple references.

As for evaluation, most existing benchmarks~\cite{oshima2025multibanana, wu2025omnigen2} adopt score-based evaluation methods similar to VIEScore~\cite{ku2024viescore}, assigning a scalar score (e.g., 0-10) for each evaluation aspect of a generated result. However, such scoring remains highly dependent on the evaluator model and often overlooks fine-grained details, such as subtle texture inconsistencies in clothing. 
Recently, several works~\cite{hu2023tifa,yarom2023you,cho2023davidsonian} have proposed checklist-based evaluation, where a set of structured questions is generated and answered by VLMs to assess specific aspects of the output, demonstrating better alignment with human judgment. To further evaluate reliability, we compare checklist-based and score-based evaluation using multiple evaluator models on MDMR-Bench. As shown in Figure~\ref{fig:inter_anno_alignment}, checklist-based evaluation achieves substantially higher inter-judge agreement indicating more consistent assessment across judges.
Building on this insight, we adopt a checklist-based evaluation protocol, where the checklist includes both global-level and atomic-level questions. Global-level questions evaluate whether the generated image satisfies the overall prompt requirements, while atomic-level questions focus on fine-grained details.
We evaluate MDMR-Bench on eleven representative open-source~\cite{wu2025qwenimagetechnicalreport, wu2025omnigen2,deng2025bagel,blackforestlabs2025flux2} and proprietary models~\cite{seedream2025seedream,seedream2026seedream5lite,nanobananapro2025, nanobanana2_2026,openai2025gptimage15}. Some findings include: current models exhibit a large gap between open-source and proprietary systems, perform substantially better on entity fusion than on infographic manipulation, and obtain lower scores on subsets with more reference images. In summary, our key contributions are as follows:

\begin{itemize}[leftmargin=1em]
    \item We introduce \textbf{\name{}}, a capability-diverse benchmark for multi-reference image generation with 2,390 cases, emphasizing broader dedicated task coverage rather than a larger number of cases.
    \item Our benchmark covers diverse challenge settings, including entity fusion, attribute transfer, multi-element extraction, infographic manipulation, and reasoning-intensive generation.
    \item To the best of our knowledge, we are the first to adopt a checklist-based evaluation protocol for multi-reference image generation, enabling fine-grained diagnosis of model capabilities and failures.
\end{itemize}

\begin{table*}[t]
    \centering
    \caption{Comparison with other benchmarks. \textcolor{green!55!black}{\ding{51}} denotes dedicated coverage, \textcolor{orange!85!black}{\(\triangle\)} denotes partial or incidental coverage, and \textcolor{red!75!black}{\ding{55}} denotes no coverage.}
    \label{tab:intro_comparision}
    \centering
    \scalebox{0.75}{
    \begingroup
\setlength{\tabcolsep}{1.2pt}
\renewcommand{\arraystretch}{1.12}
\newcommand{\cmark}{\textcolor{green!55!black}{\ding{51}}}
\newcommand{\pmark}{\textcolor{orange!85!black}{\(\triangle\)}}
\newcommand{\xmark}{\textcolor{red!75!black}{\ding{55}}}

\begin{threeparttable}
\centering
\begin{tabular}{lcccccccccc} 
\toprule
\textbf{Benchmark} &
\textbf{Size} &
\shortstack{\textbf{Entity}\\\textbf{Fusion}} &
\shortstack{\textbf{Multi-Element}\\\textbf{Extraction}} &
\shortstack{\textbf{Attribute}\\\textbf{Transfer}} &
\shortstack{\textbf{Infographic}\\\textbf{Manip.}} &
\shortstack{\textbf{Visual}\\\textbf{Reasoning}} &
\shortstack{\textbf{Global}\\\textbf{Eval.}} &
\shortstack{\textbf{Atomic}\\\textbf{Eval.}} &
\textbf{Metrics} \\
\midrule
DreamBooth~\cite{ruiz2023dreambooth}        & 75    & \cmark & \xmark & \pmark & \xmark & \xmark & \cmark & \xmark & CLIP Fidelity     \\
DreamOmni2~\cite{xia2025dreamomni2}         & 319   & \cmark & \xmark & \cmark & \pmark & \xmark & \cmark & \xmark & Score-based       \\
MICo-Bench~\cite{wei2025mico}               & 897   & \cmark & \cmark & \xmark & \xmark & \xmark & \cmark & \xmark & Score-based       \\
MultiBanana~\cite{oshima2025multibanana}    & 3769  & \cmark & \xmark & \cmark & \pmark & \xmark & \cmark & \xmark & Score-based       \\
VTEdit-Bench~\cite{liang2026vtedit}         & 24220 & \cmark & \xmark & \pmark & \xmark & \xmark & \cmark & \xmark & Score-based       \\ 
\midrule
\name{} (Ours)                                & 2390  & \cmark & \cmark & \cmark & \cmark & \cmark & \cmark & \cmark & Checklist-based   \\
\bottomrule
\end{tabular}
\end{threeparttable}

\endgroup}
    \label{tab:comparison}
    \vspace{-3ex}
\end{table*}

\section{Related Works}
\subsection{From Text and Single-Reference to Multi-Reference Image Generation}
Text-to-image generation has achieved remarkable progress since the emergence of stable diffusion~\cite{Rombach2022stablediffusion}.
Subsequent works such as Imagen~\cite{saharia2022photorealistic}, Flux~\cite{blackflux2024}, Qwen-Image~\cite{wu2025qwenimagetechnicalreport} and Z-Image~\cite{team2025zimage} have significantly improved generation quality and text alignment. Building on these advances, research has further explored single-reference image generation. These approaches mainly fall into two categories: subject-driven generation, which preserves the identity of a reference subject (e.g., DreamBooth~\cite{ruiz2023dreambooth} and OmniControl~\cite{xie2023omnicontrol}), and editing-based generation, which modifies an input image according to instructions (e.g., SeedEdit3~\cite{wang2025seededit30fasthighquality}, FLUX-Kontext~\cite{blackforestlabs2025kontext}, Emu-Edit~\cite{sheynin2024emu}, and DreamOmni~\cite{xia2025dreamomni}). However, textual instructions alone are often insufficient to precisely describe complex visual attributes, such as hairstyles, poses, artistic styles, or specific identities. To address this limitation, recent works have begun to incorporate multiple reference images as complementary visual guidance.
Representative models include Qwen-Image-Edit~\cite{wu2025qwenimagetechnicalreport}, DreamOmni2~\cite{xia2025dreamomni2}, the GPT-Image series~\cite{openai2025gptimage, openai2025gptimage15}, the NanoBanana series~\cite{nanobanana2_2026,nanobananapro2025}, and the Seedream series~\cite{seedream2025seedream,seedream2026seedream5lite}.
However, it remains unclear how well these models perform on multi-reference image generation, due to the lack of reliable and fine-grained evaluation benchmarks.

\subsection{Benchmarks for Reference-Based Image Generation}
Existing benchmarks can be broadly categorized into single-reference and multi-reference settings.
\vspace{-0.6em}
\paragraph{Single-Reference Benchmarks}
Early benchmarks such as MagicBrush~\cite{zhang2023magicbrush} and EMU-Edit~\cite{sheynin2024emu} introduce large-scale, instruction-driven image editing datasets covering diverse tasks, but suffer from relatively low-quality input images and limited coverage of task types.
Subsequent works~\cite{ye2025imgedit,yu2025anyedit, liu2025step1x-edit,ma2024i2ebench, jiang2026geditv2} focus on improving realism and coverage. For example, GEdit-Bench~\cite{liu2025step1x-edit} proposes a general-purpose benchmark grounded in real user instructions, while its follow-up version GEdit-v2~\cite{jiang2026geditv2} further improves human-centric coverage and consistency evaluation. Reason-Edit~\cite{huang2024smartedit} focuses on reasoning-intensive instruction-based image editing, which we also incorporate as a visual reasoning dimension in our multi-reference benchmark.
\vspace{-0.6em}
\paragraph{Multi-Reference Benchmarks}
OmniContext~\cite{wu2025omnigen2} introduces a multi-reference benchmark with around 400 cases focusing on scene, character, and object fusion, while MICo-Bench~\cite{wei2025mico} covers 897 cases across object-centric, human-centric, human-object interaction, and De\&Re tasks; however, both lack coverage of fine-grained attribute transfer. DreamOmni2-Bench~\cite{xia2025dreamomni2} convers a relatively comprehensive set of tasks, but contains only 319 cases, limiting its suitability for large-scale and fine-grained evaluation, and lacks a systematic evaluation protocol.  MultiBanana~\cite{oshima2025multibanana} extends the number of reference images up to eight and systematically explores variations such as domain mismatch (e.g., photo vs.\ anime), scale mismatch, rare concepts. However, it primarily varies the properties of input references and lacks explicit task-level difficulty modeling. In addition, its visual styles are relatively homogeneous, limiting the realism and diversity of evaluation scenarios.

\subsection{Evaluation Metrics and Protocols}
Many early single-image editing benchmarks~\cite{yu2025anyedit,zhang2023magicbrush,sheynin2024emu, huang2024smartedit} provide ground-truth images, enabling the use of similarity-based metrics such as L1 distance, CLIP~\cite{hessel2021clipscore}, and DINO~\cite{caron2021emerging} for evaluation. However, the target output is often not uniquely defined. As VLMs become increasingly powerful, evaluation can move beyond traditional similarity-based metrics toward more semantic and instruction-aware assessment. Recent works adopt VLM-based evaluators with score-based approaches, such as VIEScore~\cite{ku2024viescore}, ImgEdit-Bench~\cite{ye2025imgedit}, and GEdit-Bench~\cite{liu2025step1x-edit}, which produce one or a few scalar scores (e.g., per-dimension ratings), yet remain coarse-grained and unstable cross-evaluation consistency. More recently, an increasingly popular remedy in T2I evaluation is checklist-based evaluation, which decomposes a prompt into atomic yes/no questions answered by VLMs.
This paradigm is pioneered by TIFA~\cite{hu2023tifa}, and some following works~\cite{lin2026aegis,li2025easier,sun2025t2i,cho2023davidsonian} have been shown to yield finer granularity, better interpretability, and stronger alignment with human judgment than holistic scoring.
However, in multi-reference image generation, where fidelity must be verified against multiple heterogeneous references simultaneously, checklist-based evaluation remains largely unexplored.
We therefore adopt a checklist-based evaluation philosophy for multi-reference image generation, which provides more fine-grained, interpretable, and stable assessment than score-based evaluation.

\section{\name{}}
\label{sec:bench_design}
In this section, we introduce \name{} as a multi-dimensional benchmark for multi-reference image generation, covering diverse and realistic generation scenarios such as entity fusion, attribute transfer, multi-element extraction, infographic manipulation, and reasoning-intensive generation.

\subsection{Challenging Dimensions}
We organize the benchmark around representative capability settings in multi-reference image generation. Entity Fusion and Attribute Transfer are established settings in prior benchmarks, while Multi-Element Extraction tests selecting several elements from one source, Infographic Manipulation tests structured content and layout editing, and Visual Reasoning tests implicit relationships. These settings cover complementary capabilities. Specifically, we construct several challenge settings spanning different aspects of multi-reference generation, as described below.

\begin{itemize}[leftmargin=1em]
    \item \textbf{Entity Fusion and Attribute Transfer.} 
    We include common multi-reference generation scenarios involving entity fusion and attribute transfer to evaluate models' ability to accurately combine identities, attributes, and visual concepts from multiple reference images. In this subset, we vary the number of reference images from 2 to 4 to further assess multi-reference composition capability.
    \item \textbf{Multi-Element Extraction.} 
    Existing benchmarks~\cite{oshima2025multibanana, xia2025dreamomni2} often assume that each reference image provides only a single relevant element. In contrast, we include scenarios where multiple relevant elements must be extracted and integrated from a single reference, evaluating models' ability to precisely localize and compose multiple visual attributes simultaneously.
    \item \textbf{Infographic Manipulation.} 
    Recent models~\cite{nanobananapro2025, openai2025gptimage15} have demonstrated strong capabilities in infographic-oriented image generation. We therefore include infographic manipulation scenarios involving visual elements such as artistic text and logos to evaluate models' ability in design-oriented editing and structured visual composition.
    \item \textbf{Visual Reasoning.} 
    We include reasoning scenarios where the required modifications are not explicitly specified in the instructions and must be inferred from references, evaluating higher-level reasoning and implicit relationship understanding across multiple inputs.
\end{itemize}

\begin{figure}[t]
\vspace{-6ex}
    \centering
    \includegraphics[width=\linewidth]{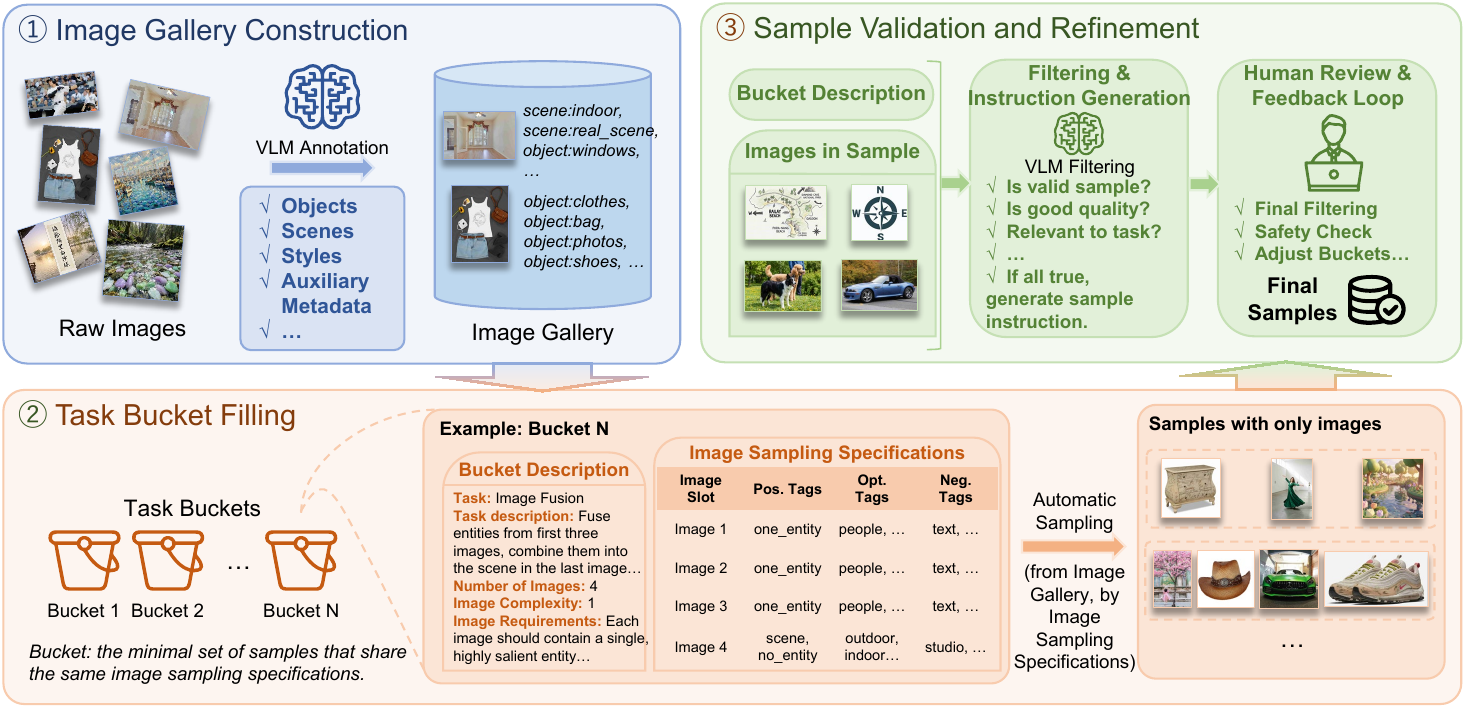}
    \caption{Sample construction pipeline of \name{}.}
    \label{fig:sample_construction}
    \vspace{-3ex}
\end{figure}

\subsection{Sample Construction}
The samples in \name{} are primarily constructed via an automatic and scalable pipeline, consisting of three stages: (1) Image Gallery Construction, (2) Task Bucket Filling, and (3) Sample Validation and Refinement, as illustrated in Figure~\ref{fig:sample_construction}. 
Although most samples are generated through this unified pipeline, certain settings require specialized construction procedures, which are introduced in their corresponding subsections.

\textbf{Image Gallery Construction}.
In the first stage, we construct a large-scale image gallery that serves as the foundation for subsequent benchmark construction. Images are primarily collected from the LAION~\cite{schuhmann2022laion5b} with resolution above 1024. To complement web-collected images, we additionally generate synthetic images using Nano Banana Pro and GPT-Image-1.5. Specifically, captions are first extracted from real images and then used as prompts for image generation, aiming to retain the depicted content while obtaining cleaner images. In the final benchmark, 48.2\% of cases contain at least one synthetic reference image, and 29\% of reference images are synthetic. A paired real-versus-synthetic experiment is reported in Appendix~\ref{app:real_synthetic}. Each image is automatically annotated using Gemini-3-Pro~\cite{comanici2025gemini}, which assigns structured semantic tags, including high-level content categories (e.g., people, animals, objects, scenes, and posters), fine-grained visual attributes (e.g., hairstyle, texture, facial expression, filter, and painting style), and auxiliary metadata such as the number of individuals, age, body framing (e.g., full-body or half-body), and facing direction. The resulting image pool contains approximately 500K images.

\textbf{Task Bucket Filling}.
In the second stage, we define a set of task buckets for sample construction.
Each bucket specifies a task type together with the corresponding reference image sampling specifications and requirements.
For example, a three-image fusion bucket requires three reference images sampled according to predefined entity and attribute constraints.
To ensure diversity within each task type (e.g., three-image fusion), we divide tasks into multiple buckets based on variations in image categories and compositions.
Each bucket is populated with 10–50 samples.
We organize sampling using two categories of annotations from the image gallery: entity tags, including person, animal, object, scene, text, outfit, and decoration, and attribute tags, including artistic style, filter, era, texture, color, hair, expression, makeup, and pose. To further improve diversity, we incorporate auxiliary tags during sampling, such as the number of people in an image and the amount of text in infographic images.

\textbf{Sample Validation and Refinement.}
In the final stage, we perform sample validation and refinement through a multi-step process. First, given the task description, we use Gemini-3-Pro~\cite{comanici2025gemini} to verify whether the sampled image combinations are valid and consistent with the intended task. Next, Gemini-3-Pro is used to generate corresponding instructions in both English and Chinese. 
We then conduct human verification to filter out low-quality or ambiguous cases, including those that violate task constraints, contain low-quality images (e.g., heavy cropping or artifacts), or result in semantically or visually inconsistent compositions.

\begin{figure*}[t]
\vspace{-6ex}
    \centering
    \begin{minipage}[t]{0.485\textwidth}
        \vspace{0pt}
        \centering
        \includegraphics[width=\linewidth]{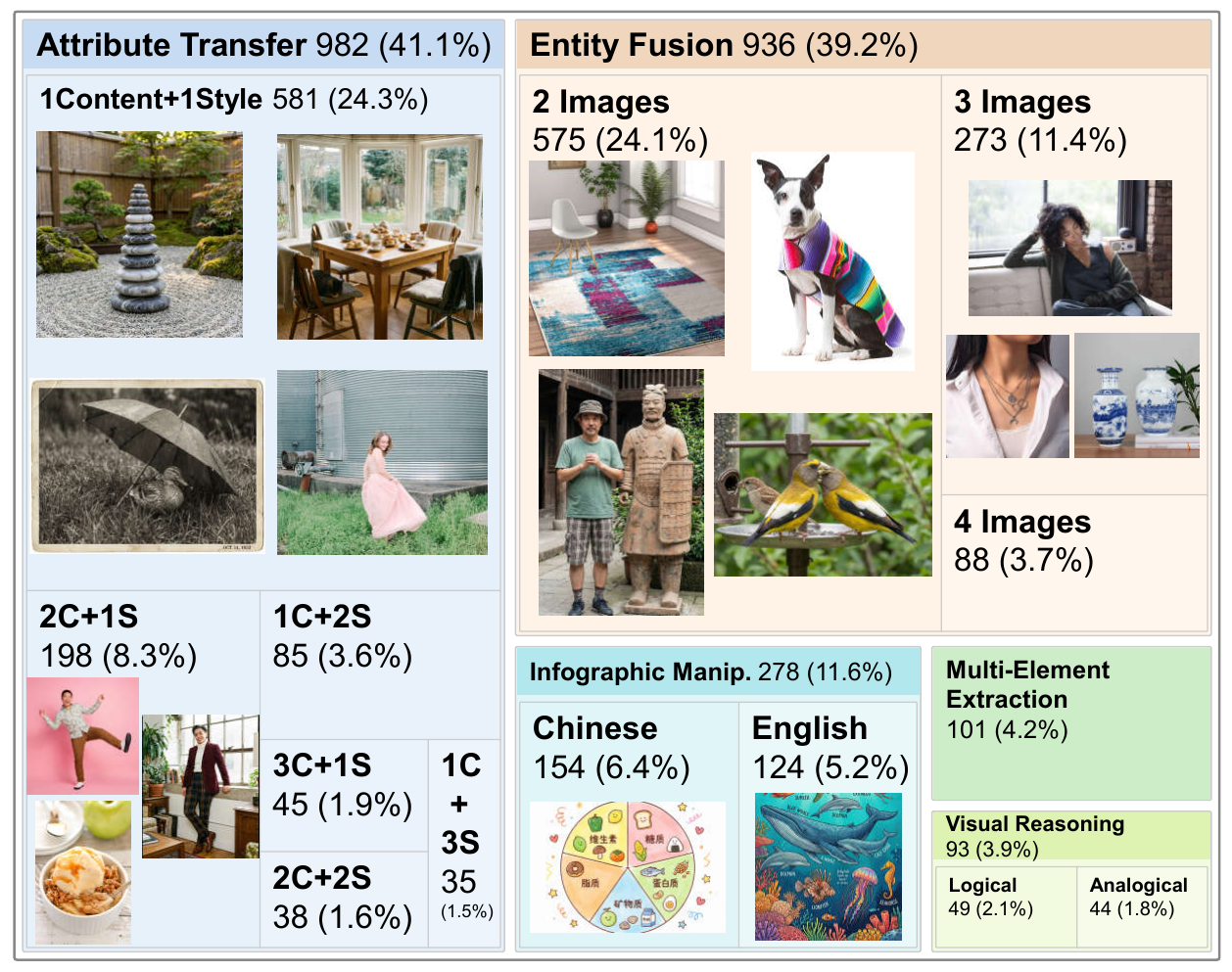}
        \subcaption{Task distribution.}\label{fig:sample_task_distribution}
    \end{minipage}
    \hspace{0.005\textwidth}
    \begin{minipage}[t]{0.485\textwidth}
        \vspace{0pt}
        \centering
        \includegraphics[width=\linewidth]{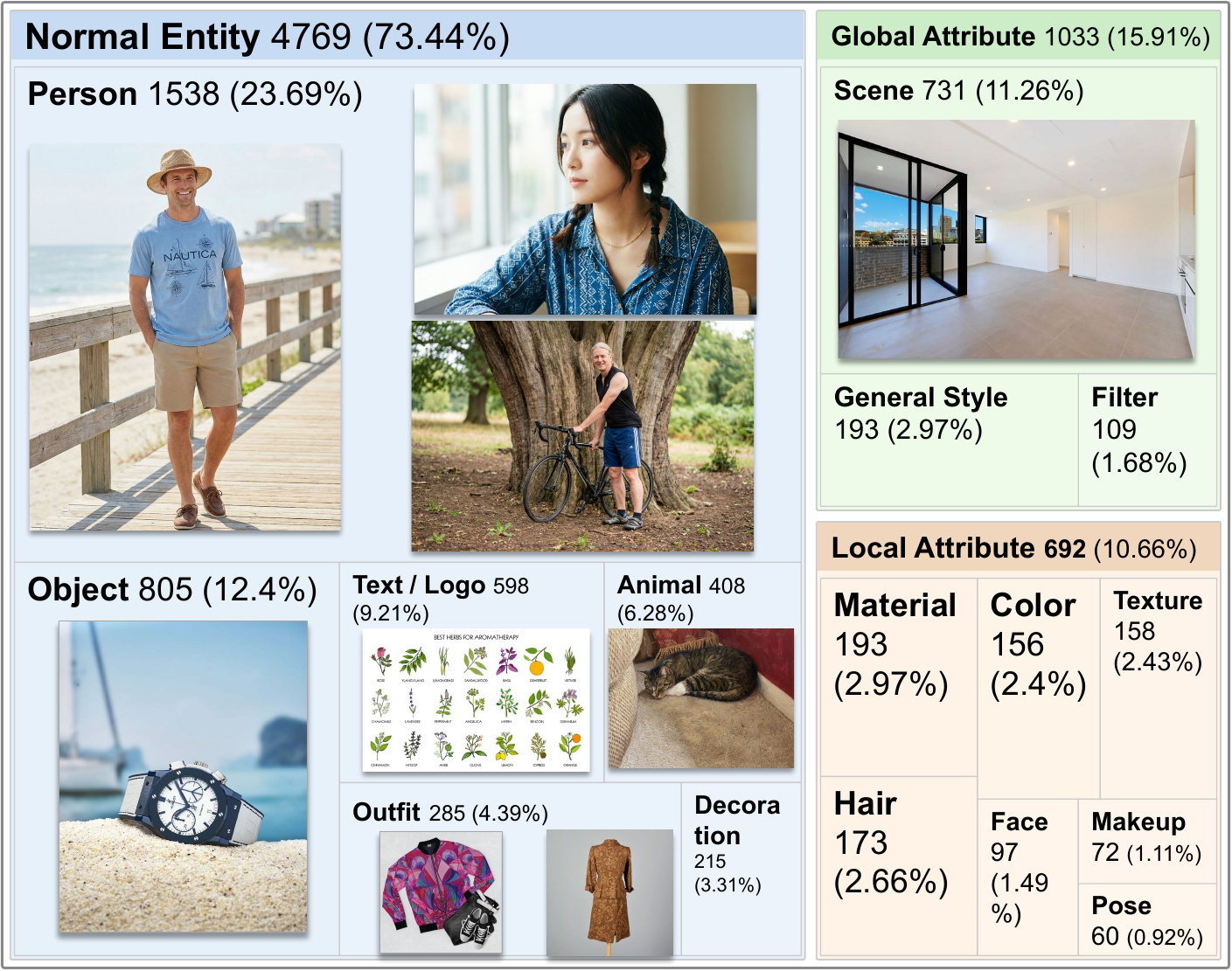}
        \subcaption{Image type distribution.}\label{fig:image_type_distribution}
    \end{minipage}
    \hspace{0.005\textwidth}
    \caption{Benchmark statistics of \name{}.}\label{fig:benchmark_statistics}
    \vspace{-3ex}
\end{figure*}

With the above pipeline, we construct 936 entity fusion samples and 982 attribute transfer samples, each containing 2 to 4 reference images. We further construct 101 multi-element extraction cases, where multiple entities or attributes are extracted from one reference image and applied to another.
For infographic manipulation, we construct 278 cases (155 Chinese and 123 English) covering operations such as addition, replacement, color modification, font change, texture/material editing, and style transfer. For addition and replacement tasks, target elements are extracted from or generated based on the original image to construct realistic and controllable editing scenarios. For visual reasoning, we construct cases spanning temporal (e.g., time and season), stylistic, state, spatial, and layout transformations, as well as operations such as addition, replacement, removal, and extraction. We first use VLMs to generate candidate reasoning formulations, which are then instantiated into concrete cases using GPT and Nano Banana Pro. The resulting cases are further divided into two categories: analogical reasoning, where reference images demonstrate the target transformation pattern, and logical reasoning, where the target output must be inferred from multiple reference images. We provide key statistics of \name{}, with additional analysis included in the appendix.
Figure~\ref{fig:sample_task_distribution} shows the distribution of different challenge settings in the benchmark. Figure~\ref{fig:image_type_distribution} reports the image type distribution, organized into normal entities, global attributes, and local attributes.

\subsection{Evaluation Metrics}
We adopt a binary checklist-based protocol rather than asking a judge model for a single holistic score. Each generated image is evaluated along five dimensions: \textbf{Following}, whether the output strictly follows the prompt instruction; \textbf{Consistency}, whether reference elements and fine-grained details are faithfully preserved; \textbf{Composition}, whether generated elements are arranged and interact in a plausible way; \textbf{Minimality}, whether the output avoids unrequested modifications; and \textbf{Quality}, whether the image is realistic, aesthetically coherent, and free of obvious artifacts.

\begin{wrapfigure}{r}{0.32\textwidth}
    \vspace{-0.8em}
    \centering
    \includegraphics[width=0.32\textwidth]{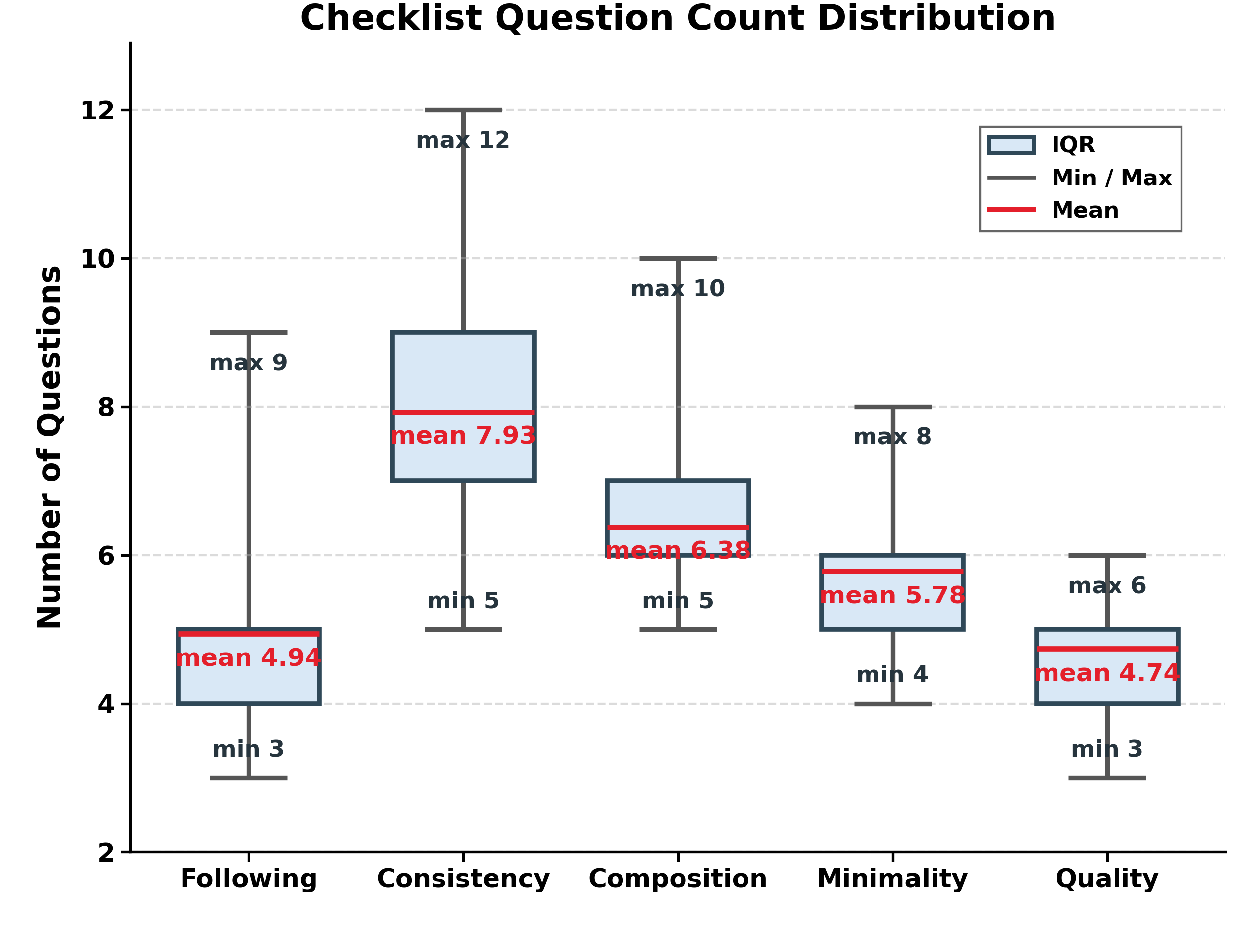}
    \caption{Checklist distribution.}
    \label{fig:checklist_question_count_stats}
    \vspace{-3.0ex}
\end{wrapfigure}

For each sample, every evaluation aspect contains two equally weighted groups of yes/no questions: \emph{global-level questions}, which evaluate overall prompt-level requirements, and \emph{atomic-level questions}, which focus on fine-grained object- and attribute-level details. These questions are automatically generated by Gemini-3-Pro based on the instruction and reference images. We manually audit a subset of the generated checklist questions; the audit protocol and results are provided in Appendix~\ref{app:checklist_audit}. Figure~\ref{fig:checklist_question_count_stats} shows the distribution of the number of questions in the checklist. During evaluation, Gemini-3-Pro serves as the default VLM judge and answers each question with ``Y'' or ``N'' and provides a short reason. We score each dimension by its yes rate and compute the item-level overall score with weights 0.20, 0.25, 0.25, 0.20, and 0.10 for Following, Consistency, Composition, Minimality, and Quality, respectively. To reduce bias from uneven subtask sizes, Entity Fusion, Visual Reasoning, Attribute Transfer, and Infographic Manipulation are first averaged within each subtask and then across subtasks; Multi-Element Extraction is averaged directly because it has no subtask split. The final score is the equal-weight mean over the five task families, using judgments from the default evaluator alone. Consequently, task sample counts do not directly determine their contribution to Overall. The dimension weights are manually specified: Consistency and Composition emphasize preserving and integrating multiple references; Following measures instruction adherence; Minimality penalizes unrequested changes; and Quality retains a basic visual-quality check with lower weight to emphasize reference-specific capabilities.

\section{Evaluation Results}
\label{sec:evaluation_results}

\subsection{Experimental Setup}
\textbf{Baseline Models}
We evaluate leading proprietary systems that support multi-reference generation, including GPT-Image-2~\cite{openai2026gptimage2}, Nano Banana 2~\cite{nanobanana2_2026}, Nano Banana Pro~\cite{nanobananapro2025}, GPT-Image-1.5~\cite{openai2025gptimage15}, Seedream 4.0~\cite{seedream2025seedream}, and Seedream 5.0-Lite~\cite{seedream2026seedream5lite}. These models represent current high-performing commercial image generation systems with strong instruction-following and multi-image conditioning capabilities.  We also include prevalent open-source baselines, including BAGEL~\cite{deng2025bagel}, DreamOmni2~\cite{xia2025dreamomni2}, OmniGen2~\cite{wu2025omnigen2}, Qwen-Image-Edit~\cite{wu2025qwenimagetechnicalreport}, and FLUX.2~\cite{blackforestlabs2025flux2}. These models cover a diverse set of unified multimodal generation and editing architectures. Evaluating them alongside proprietary systems allows us to compare the capability gap between closed and open models, and to identify which difficulty dimensions remain challenging for the broader research community.

\textbf{Output Resolution.}
Each sample is annotated with a target output resolution. During evaluation, we select the closest supported resolution for each model according to its native resolution settings, using the same selection rule across systems.

\subsection{Experimental Results}

\begin{table}[t]
\centering
\small
\setlength{\tabcolsep}{3.2pt}
\caption{Primary leaderboard under the default Gemini-3-Pro judge, in percent. GPT-5.4 results are reported separately in Appendix~\ref{app:cross_judge}.}
\label{tab:main_results_placeholder}
\begin{tabular}{lrrrrrr}
\toprule
Model & Following & Consistency & Composition & Minimality & Quality & Overall \\
\midrule
\rowcolor{gray!15}\multicolumn{7}{c}{\textbf{Closed-source Models}} \\
GPT-Image-2~\cite{openai2026gptimage2} & 90.1 & 87.4 & 90.8 & 84.5 & 93.2 & 88.8 \\
Nano Banana Pro~\cite{nanobananapro2025} & 91.2 & 84.6 & 83.9 & 80.4 & 87.0 & 85.1 \\
Nano Banana 2~\cite{nanobanana2_2026} & 88.0 & 82.0 & 81.2 & 72.1 & 84.1 & 81.2 \\
Seedream 5.0~\cite{seedream2026seedream5lite} & 86.2 & 74.0 & 81.1 & 71.5 & 84.3 & 78.8 \\
Seedream 4.0~\cite{seedream2025seedream} & 85.3 & 73.4 & 80.7 & 69.7 & 82.8 & 77.8 \\
GPT-Image-1.5~\cite{openai2025gptimage15} & 90.3 & 72.1 & 85.9 & 56.8 & 85.6 & 77.5 \\
\midrule
\rowcolor{gray!15}\multicolumn{7}{c}{\textbf{Open-source Models}} \\
FLUX.2~\cite{blackforestlabs2025flux2} & 46.5 & 50.0 & 51.1 & 51.5 & 63.8 & 51.3 \\
Qwen-Image-Edit~\cite{wu2025qwenimagetechnicalreport} & 47.7 & 46.4 & 51.5 & 49.1 & 61.3 & 49.9 \\
BAGEL~\cite{deng2025bagel} & 37.3 & 40.7 & 34.7 & 58.1 & 39.8 & 41.9 \\
DreamOmni2~\cite{xia2025dreamomni2} & 43.9 & 31.0 & 49.8 & 33.1 & 63.0 & 41.9 \\
OmniGen2~\cite{wu2025omnigen2} & 30.1 & 22.2 & 34.6 & 21.1 & 42.1 & 28.6 \\
\bottomrule
\end{tabular}
\end{table}

Table~\ref{tab:main_results_placeholder} reports scores from Gemini-3-Pro, the default evaluator. GPT-Image-2 ranks first (88.79), followed by Nano Banana Pro (85.1) and Nano Banana 2 (81.2). FLUX.2 leads the open-source models (51.3). Proprietary models outperform open-source models across all five dimensions. GPT-5.4 results are reported separately as a robustness analysis in Appendix~\ref{app:cross_judge}.

The dimension-level results further reveal different failure modes. Proprietary models generally obtain high Following and Quality scores, but their Consistency, Composition, and especially Minimality scores are lower, suggesting that even strong models may copy the wrong details, compose reference elements imperfectly, or introduce unintended changes. For open-source models, Quality is often higher than Following and Consistency, indicating that they can generate visually plausible images but frequently fail to satisfy the precise multi-reference constraints. These results show that \name{} does not merely rank models by image realism, but also exposes whether models can faithfully bind, preserve, and compose information from multiple references.

\subsection{Task-Level Results}

\begin{table}[t]
\centering
\small
\setlength{\tabcolsep}{3.0pt}
\caption{Task-level results for eleven models under Gemini-3-Pro. Scores are percentages.}
\label{tab:task_type_results}
\begin{tabular}{lcccccc}
\toprule
\textbf{Model} &
\shortstack{\textbf{Entity}\\\textbf{Fusion}} &
\shortstack{\textbf{Multi-Element}\\\textbf{Extraction}} &
\shortstack{\textbf{Visual}\\\textbf{Reasoning}} &
\shortstack{\textbf{Attribute}\\\textbf{Transfer}} &
\shortstack{\textbf{Infographic}\\\textbf{Manip.}} &
\textbf{Overall} \\
\midrule
\rowcolor{gray!15}\multicolumn{7}{c}{\textbf{Closed-source Models}} \\
GPT-Image-2~\cite{openai2026gptimage2} & 91.5 & 89.6 & 84.8 & 90.2 & 87.6 & 88.7 \\
Nano Banana Pro~\cite{nanobananapro2025} & 84.9 & 87.2 & 85.6 & 83.3 & 84.7 & 85.1 \\
Nano Banana 2~\cite{nanobanana2_2026} & 82.5 & 85.6 & 75.9 & 82.4 & 79.8 & 81.2 \\
Seedream 5.0~\cite{seedream2026seedream5lite} & 77.9 & 81.3 & 83.6 & 78.6 & 72.5 & 78.8 \\
Seedream 4.0~\cite{seedream2025seedream} & 77.9 & 83.1 & 83.7 & 76.6 & 67.9 & 77.8 \\
GPT-Image-1.5~\cite{openai2025gptimage15} & 81.8 & 78.4 & 76.5 & 82.3 & 68.5 & 77.5 \\
\midrule
\rowcolor{gray!15}\multicolumn{7}{c}{\textbf{Open-source Models}} \\
FLUX.2~\cite{blackforestlabs2025flux2} & 73.5 & 54.1 & 36.7 & 47.9 & 44.0 & 51.3 \\
Qwen-Image-Edit~\cite{wu2025qwenimagetechnicalreport} & 66.9 & 55.3 & 44.3 & 44.1 & 39.2 & 49.9 \\
BAGEL~\cite{deng2025bagel} & 41.5 & 39.9 & 52.1 & 35.6 & 40.5 & 41.9 \\
DreamOmni2~\cite{xia2025dreamomni2} & 50.0 & 44.2 & 34.2 & 43.1 & 38.0 & 41.9 \\
OmniGen2~\cite{wu2025omnigen2} & 44.9 & 29.0 & 22.7 & 26.9 & 19.7 & 28.6 \\
\bottomrule
\end{tabular}
\vspace{-3ex}
\end{table}

Table~\ref{tab:task_type_results} breaks down performance by task type and reveals that different tasks stress different capabilities. We have the following findings: 1) Proprietary models exhibit relatively balanced performance across all task types, while open-source models show substantial performance variation across different settings. This suggests that current proprietary systems have developed more comprehensive multi-reference generation capabilities, whereas open-source models still rely more heavily on task-specific strengths. 2)  Interestingly, Multi-Element Extraction, which can be viewed as a special case of Entity Fusion and Attribute Transfer, is often easier for current models than using multiple separate reference images. A possible explanation is that consolidating multiple reference elements within a single image reduces the effective multi-reference complexity, whereas increasing the number of reference images introduces longer visual token sequences and greater compositional difficulty, leading to degraded model performance. 3) Except for Nano Banana Pro, all models achieve noticeably lower performance on the Infographic Manipulation subset, suggesting that precise infographic editing remains a major challenge for current multi-reference image generation systems. 4) For reasoning-intensive generation, models with stronger vision-language understanding capabilities achieve substantially better performance. In particular, Nano Banana Pro and Seedream 4 outperform most other systems on reasoning-related tasks. Interestingly, BAGEL also attains relatively strong reasoning performance despite weaker overall image quality. As a unified mixture-of-transformers architecture that jointly models visual understanding and generation, BAGEL suggests that effective multi-reference reasoning depends heavily on integrated visual comprehension capabilities.

\subsection{Reliability of VLM Judge}
\label{sec:vlm_judger}
To validate our checklist-based evaluation protocol, we compare VLM judge decisions with human judgments. As shown in Table~\ref{tab:judge_reliability}, the three proprietary judges achieve similar agreement ratios with humans, with Gemini-3-Pro~\cite{comanici2025gemini}, GPT-5.4~\cite{openai2026gpt54}, and Doubao-1.8~\cite{bytedanceseed2025seed18} reaching 0.853, 0.854, and 0.849, respectively. Their rank correlations are also strong, especially Gemini-3-Pro~\cite{comanici2025gemini}, which obtains the highest Spearman correlation (0.862) and Cohen's $\kappa$ (0.642). These results support the use of VLM judges as scalable proxies for human evaluation in \name{}.

\begin{table}[t]
\centering
\small
\setlength{\tabcolsep}{8.0pt}
\caption{Correlation between human and VLM judges.}
\label{tab:judge_reliability}
\begin{tabular}{lcccc}
\toprule
\textbf{Judge} &
\textbf{Agreement Ratio} &
\textbf{Pearson $r$} &
\textbf{Spearman $r$} &
\textbf{Cohen's $\kappa$} \\
\midrule
Gemini-3-Pro~\cite{comanici2025gemini} & 0.853 & 0.643 & 0.862 & 0.642 \\
GPT-5.4~\cite{openai2026gpt54} & 0.854 & 0.621 & 0.834 & 0.618 \\
Doubao-1.8~\cite{bytedanceseed2025seed18} & 0.849 & 0.617 & 0.784 & 0.617 \\
Qwen3-VL-32B~\cite{bai2025qwen3vl} & 0.800 & 0.475 & 0.725 & 0.472 \\
\bottomrule
\end{tabular}
\vspace{-2ex}
\end{table}

The human study comprises 5,924 checklist decisions from 200 benchmark cases and 800 outputs from Nano Banana Pro, Seedream 5.0, FLUX.2, and Qwen-Image-Edit. Five annotators independently answer Y/N while viewing the instruction, all references, and the generated output. Generator identities are hidden and output order is randomized; final human labels use majority voting. Human inter-annotator Krippendorff's $\alpha$ is 0.88. Task counts and further protocol details are given in Appendix~\ref{app:human_protocol}.

Among the evaluated judges, Gemini-3-Pro~\cite{comanici2025gemini} provides the most balanced reliability across agreement ratio, Pearson correlation, Spearman correlation, and Cohen's $\kappa$, and serves as the default evaluator for the primary leaderboard and diagnostic analyses. Qwen3-VL-32B~\cite{bai2025qwen3vl} is less aligned with human ratings, but still achieves a reasonable agreement ratio of 0.800, suggesting that open-source VLMs can serve as useful alternatives when proprietary APIs are unavailable.

\subsection{Diagnostic Analysis}
We conduct two diagnostic analyses on the Attribute Transfer subset: (1) the impact of different reference content types, and (2) observed performance across subsets with different numbers of input images.
\begin{figure*}[t]
  \centering
  \includegraphics[width=\textwidth]{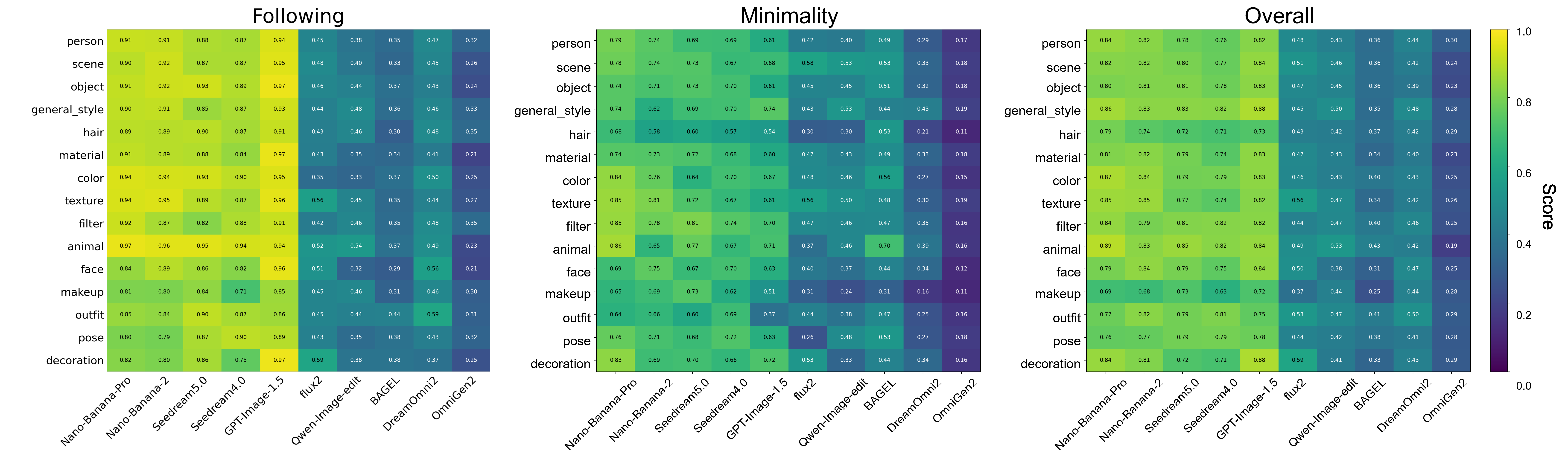}
  \caption{Diagnostic heatmaps for the original ten models under Gemini-3-Pro.}\label{fig:style_transfer_diagnostic_heatmaps}
  \vspace{-1ex}
\end{figure*}

\begin{figure*}[t]
  \centering
  \includegraphics[width=\textwidth]{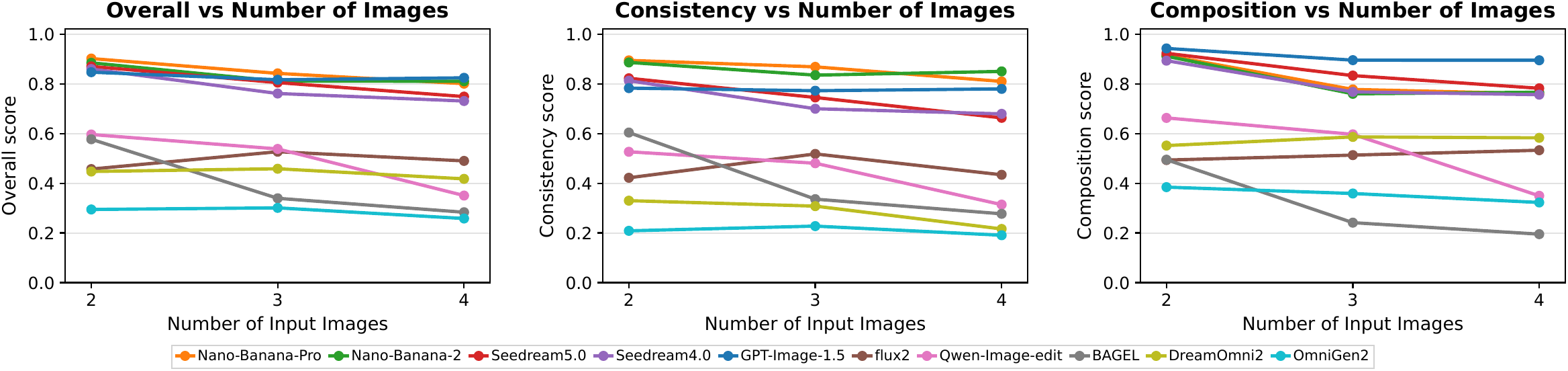}
  \caption{Diagnostic curves for the original ten models under Gemini-3-Pro across different numbers of images. Overall measures aggregate performance, Consistency measures reference fidelity, and Composition measures multi-reference integration.}\label{fig:num_images_diagnostic_curves}
  \vspace{-3ex}
\end{figure*}

\textbf{Impact of Reference Content Types.} We observe as shown in Figure~\ref{fig:style_transfer_diagnostic_heatmaps} that current models perform substantially worse on human-related attribute transfer tasks, such as facial expression, makeup, outfit, pose, and hairstyle, likely because these tasks require precise manipulation of human appearance and identity.
The dimension-wise heatmaps also reveal trade-offs hidden by the aggregate score. For example, GPT-Image-1.5 achieves strong Following, indicating that it can satisfy the requested attribute-transfer instructions across content types, but this comes with a much lower Minimality score. In other words, it often follows the requested transfer by changing content that should have remained untouched, which limits its Overall score.

\textbf{Impact of the Number of Reference Images.}
Figure~\ref{fig:num_images_diagnostic_curves} shows lower performance on Attribute Transfer subsets containing more reference images. From two to four references, average Overall is lower by roughly 0.106, and Composition by roughly 0.122, on the plotted 0--1 scale.

\section{Conclusion}
We introduced \textbf{\name{}}, a 2,390-case, multi-dimensional benchmark for evaluating multi-reference image generation. \name{} covers realistic multi-reference scenarios across entity fusion, attribute transfer, multi-element extraction, infographic manipulation, and visual reasoning-aware generation. To move beyond coarse holistic scores, we further adopt a checklist-based evaluation protocol that decomposes each generation result into interpretable and visually verifiable judgments.
GPT-Image-2 leads the primary leaderboard evaluated by Gemini-3-Pro. Our evaluation of representative proprietary and open-source models shows that multi-reference image generation remains challenging. Proprietary systems substantially outperform open-source models, but even strong models exhibit non-trivial failures in reference consistency, composition, and minimality. The task-level results further show that models are relatively stronger at visible entity fusion, while infographic manipulation, visual reasoning-aware generation, and fine-grained attribute transfer remain difficult. Diagnostic analyses reveal that subsets with more references have lower performance, and that instruction following can come at the cost of unintended modifications. We hope \name{} provides a useful foundation for measuring these capability gaps and guiding future progress toward more faithful, controllable, and reliable multi-reference generation systems.

\ifpublicversion
  \section*{Dataset and Code Availability}
  The \name{} dataset is publicly available at \url{\dataseturl}. The evaluation code and release assets are publicly available at \url{\codeurl}.
\fi
\bibliographystyle{plain}
\bibliography{\sharedroot/bibliography/references}

\begin{thebibliography}{10}

\bibitem{bai2025qwen3vl}
Shuai Bai, Yuxuan Cai, Ruizhe Chen, Keqin Chen, Xionghui Chen, Zesen Cheng,
  Lianghao Deng, Wei Ding, Chang Gao, Chunjiang Ge, et~al.
\newblock Qwen3-vl technical report.
\newblock {\em arXiv preprint arXiv:2511.21631}, 2025.

\bibitem{blackforestlabs2025kontext}
{Black Forest Labs}.
\newblock Flux.1-kontext-dev.
\newblock 2025.
\newblock \url{https://blackforestlabs.ai/flux-1-kontext/}.

\bibitem{blackforestlabs2025flux2}
{Black Forest Labs}.
\newblock {FLUX.2: Frontier Visual Intelligence}.
\newblock \url{https://bfl.ai/blog/flux-2}, November 2025.
\newblock Accessed: 2026-04-30.

\bibitem{bytedanceseed2025seed18}
{ByteDance Seed}.
\newblock {Seed1.8}.
\newblock \url{https://seed.bytedance.com/en/seed1_8}, 2025.
\newblock Accessed: 2026-05-07.

\bibitem{seedream2026seedream5lite}
{ByteDance Seed}.
\newblock {Seedream 5.0 Lite}.
\newblock \url{https://seed.bytedance.com/en/seedream5_0_lite}, 2026.
\newblock Accessed: 2026-04-30.

\bibitem{caron2021emerging}
Mathilde Caron, Hugo Touvron, Ishan Misra, Herv{\'e} J{\'e}gou, Julien Mairal,
  Piotr Bojanowski, and Armand Joulin.
\newblock Emerging properties in self-supervised vision transformers.
\newblock In {\em Proceedings of the IEEE/CVF international conference on
  computer vision}, pages 9650--9660, 2021.

\bibitem{chen2025blip3o}
Jiuhai Chen, Zhiyang Xu, Xichen Pan, Yushi Hu, Can Qin, Tom Goldstein, Lifu
  Huang, Tianyi Zhou, Saining Xie, Silvio Savarese, et~al.
\newblock Blip3-o: A family of fully open unified multimodal
  models-architecture, training and dataset.
\newblock {\em arXiv preprint arXiv:2505.09568}, 2025.

\bibitem{cho2023davidsonian}
Jaemin Cho, Yushi Hu, Roopal Garg, Peter Anderson, Ranjay Krishna, Jason
  Baldridge, Mohit Bansal, Jordi Pont-Tuset, and Su~Wang.
\newblock Davidsonian scene graph: Improving reliability in fine-grained
  evaluation for text-to-image generation.
\newblock {\em arXiv preprint arXiv:2310.18235}, 2023.

\bibitem{comanici2025gemini}
Gheorghe Comanici, Eric Bieber, Mike Schaekermann, Ice Pasupat, Noveen
  Sachdeva, Inderjit Dhillon, Marcel Blistein, Ori Ram, Dan Zhang, Evan Rosen,
  et~al.
\newblock Gemini 2.5: Pushing the frontier with advanced reasoning,
  multimodality, long context, and next generation agentic capabilities.
\newblock {\em arXiv preprint arXiv:2507.06261}, 2025.

\bibitem{deng2025bagel}
Chaorui Deng, Deyao Zhu, Kunchang Li, Chenhui Gou, Feng Li, Zeyu Wang, Shu
  Zhong, Weihao Yu, Xiaonan Nie, Ziang Song, et~al.
\newblock Emerging properties in unified multimodal pretraining.
\newblock {\em arXiv preprint arXiv:2505.14683}, 2025.

\bibitem{esser2024sd3}
Patrick Esser, Sumith Kulal, Andreas Blattmann, Rahim Entezari, Jonas Müller,
  Harry Saini, Yam Levi, Dominik Lorenz, Axel Sauer, Frederic Boesel, Dustin
  Podell, Tim Dockhorn, Zion English, Kyle Lacey, Alex Goodwin, Yannik Marek,
  and Robin Rombach.
\newblock Scaling rectified flow transformers for high-resolution image
  synthesis, 2024.

\bibitem{ghosh2023geneval}
Dhruba Ghosh, Hannaneh Hajishirzi, and Ludwig Schmidt.
\newblock Geneval: An object-focused framework for evaluating text-to-image
  alignment.
\newblock {\em Advances in Neural Information Processing Systems},
  36:52132--52152, 2023.

\bibitem{nanobananapro2025}
{Google DeepMind}.
\newblock Gemini 3 pro image (nano banana pro).
\newblock \url{https://blog.google/technology/ai/nano-banana-pro}, November
  2025.
\newblock Accessed: 2026-04-30.

\bibitem{nanobanana2_2026}
{Google DeepMind}.
\newblock Gemini 3.1 flash image (nano banana 2).
\newblock
  \url{https://blog.google/innovation-and-ai/technology/ai/nano-banana-2},
  February 2026.
\newblock Accessed: 2026-04-30.

\bibitem{hessel2021clipscore}
Jack Hessel, Ari Holtzman, Maxwell Forbes, Ronan Le~Bras, and Yejin Choi.
\newblock Clipscore: A reference-free evaluation metric for image captioning.
\newblock In {\em Proceedings of the 2021 conference on empirical methods in
  natural language processing}, pages 7514--7528, 2021.

\bibitem{hu2024dpgbench}
Xiwei Hu, Rui Wang, Yixiao Fang, Bin Fu, Pei Cheng, and Gang Yu.
\newblock Ella: Equip diffusion models with llm for enhanced semantic
  alignment.
\newblock {\em arXiv preprint arXiv:2403.05135}, 2024.

\bibitem{hu2023tifa}
Yushi Hu, Benlin Liu, Jungo Kasai, Yizhong Wang, Mari Ostendorf, Ranjay
  Krishna, and Noah~A Smith.
\newblock Tifa: Accurate and interpretable text-to-image faithfulness
  evaluation with question answering.
\newblock {\em arXiv preprint arXiv:2303.11897}, 2023.

\bibitem{huang2024smartedit}
Yuzhou Huang, Liangbin Xie, Xintao Wang, Ziyang Yuan, Xiaodong Cun, Yixiao Ge,
  Jiantao Zhou, Chao Dong, Rui Huang, Ruimao Zhang, et~al.
\newblock Smartedit: Exploring complex instruction-based image editing with
  multimodal large language models.
\newblock In {\em Proceedings of the IEEE/CVF Conference on Computer Vision and
  Pattern Recognition}, pages 8362--8371, 2024.

\bibitem{jiang2026geditv2}
Zhangqi Jiang, Zheng Sun, Xianfang Zeng, Yufeng Yang, Xuanyang Zhang, Yongliang
  Wu, Wei Cheng, Gang Yu, Xu~Yang, and Bihan Wen.
\newblock Geditbench v2: A human-aligned benchmark for general image editing,
  2026.

\bibitem{kamath2025geneval2}
Amita Kamath, Kai-Wei Chang, Ranjay Krishna, Luke Zettlemoyer, Yushi Hu, and
  Marjan Ghazvininejad.
\newblock Geneval 2: Addressing benchmark drift in text-to-image evaluation.
\newblock {\em arXiv preprint arXiv:2512.16853}, 2025.

\bibitem{krippendorff1970bivariate}
Klaus Krippendorff.
\newblock Bivariate agreement coefficients for reliability data.
\newblock In Edgar~F. Borgatta and George~W. Bohrnstedt, editors, {\em
  Sociological Methodology 1970}, volume~2, pages 139--150. Jossey-Bass, San
  Francisco, CA, 1970.

\bibitem{ku2024viescore}
Max Ku, Dongfu Jiang, Cong Wei, Xiang Yue, and Wenhu Chen.
\newblock Viescore: Towards explainable metrics for conditional image synthesis
  evaluation.
\newblock In {\em Proceedings of the 62nd Annual Meeting of the Association for
  Computational Linguistics (Volume 1: Long Papers)}, pages 12268--12290, 2024.

\bibitem{blackflux2024}
Black~Forest Labs.
\newblock Flux.
\newblock \url{https://github.com/black-forest-labs/flux}, 2024.

\bibitem{li2025easier}
Ouxiang Li, Yuan Wang, Xinting Hu, Huijuan Huang, Rui Chen, Jiarong Ou, Xin
  Tao, Pengfei Wan, Xiaojuan Qi, and Fuli Feng.
\newblock Easier painting than thinking: Can text-to-image models set the
  stage, but not direct the play?
\newblock {\em arXiv preprint arXiv:2509.03516}, 2025.

\bibitem{liang2026vtedit}
Xiaoye Liang, Zhiyuan Qu, Mingye Zou, Jiaxin Liu, Lai Jiang, Mai Xu, and Yiheng
  Zhu.
\newblock Vtedit-bench: A comprehensive benchmark for multi-reference image
  editing models in virtual try-on.
\newblock {\em arXiv preprint arXiv:2603.11734}, 2026.

\bibitem{lin2026aegis}
Jintao Lin, Bowen Dong, Weikang Shi, Chenyang Lei, Suiyun Zhang, Rui Liu, and
  Xihui Liu.
\newblock Aegis: Exploring the limit of world knowledge capabilities for
  unified mulitmodal models.
\newblock {\em arXiv preprint arXiv:2601.00561}, 2026.

\bibitem{liu2025step1x-edit}
Shiyu Liu, Yucheng Han, Peng Xing, Fukun Yin, Rui Wang, Wei Cheng, Jiaqi Liao,
  Yingming Wang, Honghao Fu, Chunrui Han, Guopeng Li, Yuang Peng, Quan Sun,
  Jingwei Wu, Yan Cai, Zheng Ge, Ranchen Ming, Lei Xia, Xianfang Zeng, Yibo
  Zhu, Binxing Jiao, Xiangyu Zhang, Gang Yu, and Daxin Jiang.
\newblock Step1x-edit: A practical framework for general image editing.
\newblock {\em arXiv preprint arXiv:2504.17761}, 2025.

\bibitem{ma2024i2ebench}
Yiwei Ma, Jiayi Ji, Ke~Ye, Weihuang Lin, Zhibin Wang, Yonghan Zheng, Qiang
  Zhou, Xiaoshuai Sun, and Rongrong Ji.
\newblock I2ebench: A comprehensive benchmark for instruction-based image
  editing.
\newblock {\em Advances in Neural Information Processing Systems},
  37:41494--41516, 2024.

\bibitem{openai2025gptimage}
{OpenAI}.
\newblock {GPT-Image-1 Model}.
\newblock \url{https://platform.openai.com/docs/models/gpt-image-1}, 2025.
\newblock Accessed: 2025-10-16.

\bibitem{openai2025gptimage15}
{OpenAI}.
\newblock {GPT-Image-1.5 Model}.
\newblock \url{https://developers.openai.com/api/docs/models/gpt-image-1.5},
  December 2025.
\newblock Accessed: 2025-12-16.

\bibitem{openai2026gpt54}
{OpenAI}.
\newblock {GPT-5.4 Model}.
\newblock \url{https://developers.openai.com/api/docs/models/gpt-5.4}, March
  2026.
\newblock Accessed: 2026-05-07.

\bibitem{openai2026gptimage2}
{OpenAI}.
\newblock {GPT-Image-2 Model}.
\newblock \url{https://developers.openai.com/api/docs/models/gpt-image-2},
  April 2026.
\newblock Accessed: 2026-05-27.

\bibitem{oshima2025multibanana}
Yuta Oshima, Daiki Miyake, Kohsei Matsutani, Yusuke Iwasawa, Masahiro Suzuki,
  Yutaka Matsuo, and Hiroki Furuta.
\newblock Multibanana: A challenging benchmark for multi-reference
  text-to-image generation.
\newblock {\em arXiv preprint arXiv:2511.22989}, 2025.

\bibitem{Rombach2022stablediffusion}
Robin Rombach, Andreas Blattmann, Dominik Lorenz, Patrick Esser, and Bj\"orn
  Ommer.
\newblock High-resolution image synthesis with latent diffusion models.
\newblock In {\em Proceedings of the IEEE/CVF Conference on Computer Vision and
  Pattern Recognition (CVPR)}, pages 10684--10695, June 2022.

\bibitem{ruiz2023dreambooth}
Nataniel Ruiz, Yuanzhen Li, Varun Jampani, Yael Pritch, Michael Rubinstein, and
  Kfir Aberman.
\newblock Dreambooth: Fine tuning text-to-image diffusion models for
  subject-driven generation.
\newblock In {\em Proceedings of the IEEE/CVF conference on computer vision and
  pattern recognition}, pages 22500--22510, 2023.

\bibitem{saharia2022photorealistic}
Chitwan Saharia, William Chan, Saurabh Saxena, Lala Li, Jay Whang, Emily~L
  Denton, Kamyar Ghasemipour, Raphael Gontijo~Lopes, Burcu Karagol~Ayan, Tim
  Salimans, et~al.
\newblock Photorealistic text-to-image diffusion models with deep language
  understanding.
\newblock {\em Advances in neural information processing systems},
  35:36479--36494, 2022.

\bibitem{schuhmann2022laion5b}
Christoph Schuhmann, Romain Beaumont, Richard Vencu, Cade Gordon, Ross
  Wightman, Mehdi Cherti, Theo Coombes, Aarush Katta, Clayton Mullis, Mitchell
  Wortsman, Patrick Schramowski, Srivatsa Kundurthy, Katherine Crowson, Ludwig
  Schmidt, Robert Kaczmarczyk, and Jenia Jitsev.
\newblock {LAION-5B}: An open large-scale dataset for training next generation
  image-text models.
\newblock In {\em Proceedings of the Neural Information Processing Systems
  Track on Datasets and Benchmarks ({NeurIPS} Datasets and Benchmarks 2022)},
  2022.

\bibitem{seedream2025seedream}
Team Seedream, Yunpeng Chen, Yu~Gao, Lixue Gong, Meng Guo, Qiushan Guo, Zhiyao
  Guo, Xiaoxia Hou, Weilin Huang, Yixuan Huang, et~al.
\newblock Seedream 4.0: Toward next-generation multimodal image generation.
\newblock {\em arXiv preprint arXiv:2509.20427}, 2025.

\bibitem{sheynin2024emu}
Shelly Sheynin, Adam Polyak, Uriel Singer, Yuval Kirstain, Amit Zohar, Oron
  Ashual, Devi Parikh, and Yaniv Taigman.
\newblock Emu edit: Precise image editing via recognition and generation tasks.
\newblock In {\em Proceedings of the IEEE/CVF Conference on Computer Vision and
  Pattern Recognition}, pages 8871--8879, 2024.

\bibitem{sun2025t2i}
Kaiyue Sun, Rongyao Fang, Chengqi Duan, Xian Liu, and Xihui Liu.
\newblock T2i-reasonbench: Benchmarking reasoning-informed text-to-image
  generation.
\newblock {\em arXiv preprint arXiv:2508.17472}, 2025.

\bibitem{LongCat-Image}
Meituan~LongCat Team, Hanghang Ma, Haoxian Tan, Jiale Huang, Junqiang Wu,
  Jun-Yan He, Lishuai Gao, Songlin Xiao, Xiaoming Wei, Xiaoqi Ma, Xunliang Cai,
  Yayong Guan, and Jie Hu.
\newblock Longcat-image technical report.
\newblock {\em arXiv preprint arXiv:2512.07584}, 2025.

\bibitem{firered2026rededit}
Super~Intelligence Team.
\newblock Firered-image-edit: A general-purpose image editing model.
\newblock 2026.

\bibitem{team2025zimage}
Z-Image Team.
\newblock Z-image: An efficient image generation foundation model with
  single-stream diffusion transformer.
\newblock {\em arXiv preprint arXiv:2511.22699}, 2025.

\bibitem{wang2025seededit30fasthighquality}
Peng Wang, Yichun Shi, Xiaochen Lian, Zhonghua Zhai, Xin Xia, Xuefeng Xiao,
  Weilin Huang, and Jianchao Yang.
\newblock Seededit 3.0: Fast and high-quality generative image editing, 2025.

\bibitem{wei2025mico}
Xinyu Wei, Kangrui Cen, Hongyang Wei, Zhen Guo, Bairui Li, Zeqing Wang, Jinrui
  Zhang, and Lei Zhang.
\newblock Mico-150k: A comprehensive dataset advancing multi-image composition.
\newblock {\em arXiv preprint arXiv:2512.07348}, 2025.

\bibitem{wu2025qwenimagetechnicalreport}
Chenfei Wu, Jiahao Li, Jingren Zhou, Junyang Lin, Kaiyuan Gao, Kun Yan, Sheng
  ming Yin, Shuai Bai, Xiao Xu, Yilei Chen, Yuxiang Chen, Zecheng Tang, Zekai
  Zhang, Zhengyi Wang, An~Yang, Bowen Yu, Chen Cheng, Dayiheng Liu, Deqing Li,
  Hang Zhang, Hao Meng, Hu~Wei, Jingyuan Ni, Kai Chen, Kuan Cao, Liang Peng,
  Lin Qu, Minggang Wu, Peng Wang, Shuting Yu, Tingkun Wen, Wensen Feng,
  Xiaoxiao Xu, Yi~Wang, Yichang Zhang, Yongqiang Zhu, Yujia Wu, Yuxuan Cai, and
  Zenan Liu.
\newblock Qwen-image technical report, 2025.

\bibitem{wu2025omnigen2}
Chenyuan Wu, Pengfei Zheng, Ruiran Yan, Shitao Xiao, Xin Luo, Yueze Wang, Wanli
  Li, Xiyan Jiang, Yexin Liu, Junjie Zhou, Ze~Liu, Ziyi Xia, Chaofan Li, Haoge
  Deng, Jiahao Wang, Kun Luo, Bo~Zhang, Defu Lian, Xinlong Wang, Zhongyuan
  Wang, Tiejun Huang, and Zheng Liu.
\newblock Omnigen2: Exploration to advanced multimodal generation.
\newblock {\em arXiv preprint arXiv:2506.18871}, 2025.

\bibitem{xia2025dreamomni2}
Bin Xia, Bohao Peng, Yuechen Zhang, Junjia Huang, Jiyang Liu, Jingyao Li, Haoru
  Tan, Sitong Wu, Chengyao Wang, Yitong Wang, et~al.
\newblock Dreamomni2: Multimodal instruction-based editing and generation.
\newblock {\em arXiv preprint arXiv:2510.06679}, 2025.

\bibitem{xia2025dreamomni}
Bin Xia, Yuechen Zhang, Jingyao Li, Chengyao Wang, Yitong Wang, Xinglong Wu,
  Bei Yu, and Jiaya Jia.
\newblock Dreamomni: Unified image generation and editing.
\newblock In {\em Proceedings of the IEEE/CVF Conference on Computer Vision and
  Pattern Recognition (CVPR)}, pages 28533--28543, June 2025.

\bibitem{xie2023omnicontrol}
Yiming Xie, Varun Jampani, Lei Zhong, Deqing Sun, and Huaizu Jiang.
\newblock Omnicontrol: Control any joint at any time for human motion
  generation.
\newblock {\em arXiv preprint arXiv:2310.08580}, 2023.

\bibitem{yarom2023you}
Michal Yarom, Yonatan Bitton, Soravit Changpinyo, Roee Aharoni, Jonathan
  Herzig, Oran Lang, Eran Ofek, and Idan Szpektor.
\newblock What you see is what you read? improving text-image alignment
  evaluation.
\newblock {\em arXiv preprint arXiv:2305.10400}, 2023.

\bibitem{ye2025imgedit}
Yang Ye, Xianyi He, Zongjian Li, Bin Lin, Shenghai Yuan, Zhiyuan Yan, Bohan
  Hou, and Li~Yuan.
\newblock Imgedit: A unified image editing dataset and benchmark.
\newblock {\em arXiv preprint arXiv:2505.20275}, 2025.

\bibitem{yu2025anyedit}
Qifan Yu, Wei Chow, Zhongqi Yue, Kaihang Pan, Yang Wu, Xiaoyang Wan, Juncheng
  Li, Siliang Tang, Hanwang Zhang, and Yueting Zhuang.
\newblock Anyedit: Mastering unified high-quality image editing for any idea,
  2025.

\bibitem{zhang2023magicbrush}
Kai Zhang, Lingbo Mo, Wenhu Chen, Huan Sun, and Yu~Su.
\newblock Magicbrush: A manually annotated dataset for instruction-guided image
  editing.
\newblock {\em Advances in Neural Information Processing Systems},
  36:31428--31449, 2023.

\end{thebibliography}
\appendix
\newpage
\section{Appendix}

\subsection{Inter-Judge Agreement Analysis Setting}
\label{appx:inter_judge}
To evaluate the reliability of different evaluation protocols, we conduct an inter-judge agreement analysis using four evaluator models: GPT-5.4~\cite{openai2026gpt54}, Doubao-Seed-1.8~\cite{bytedanceseed2025seed18}, Qwen3-VL-32B~\cite{bai2025qwen3vl}, and Gemini-3-Pro~\cite{comanici2025gemini}. We apply these evaluator models to assess outputs generated by four representative image generation systems, including Nano Banana Pro~\cite{nanobananapro2025}, FLUX.2~\cite{blackforestlabs2025flux2}, Seedream5~\cite{seedream2026seedream5lite}, and Qwen-Image-Edit~\cite{wu2025qwenimagetechnicalreport}.
For score-based evaluation, we adopt a scoring prompt similar to the evaluation protocol used in MultiBanana~\cite{oshima2025multibanana}, where evaluator models assign scalar scores for each evaluation aspect. For our checklist-based evaluation, evaluator models answer structured global-level and atomic-level questions. We compute Krippendorff's alpha~\cite{krippendorff1970bivariate} to measure inter-judge agreement across different evaluator models.
As shown in Figure~\ref{fig:inter_anno_alignment}, checklist-based evaluation consistently achieves substantially higher agreement than conventional score-based evaluation across most evaluation aspects, indicating improved evaluation consistency across judges. Separately, Gemini score-based evaluation achieves Krippendorff's $\alpha=0.5896$ against human scores using the interval metric.

\subsection{Cross-Judge Evaluation}
\label{app:cross_judge}
\label{sec:cross_judge}
The primary leaderboard in Table~\ref{tab:main_results_placeholder} uses Gemini-3-Pro alone.
However, evaluator independence is important. Therefore we conducted an additional non-Gemini judge analysis to directly test whether our conclusions depend on the Gemini judge. This supplementary analysis is not required to compute the benchmark score. Benchmark cases, generated outputs, Gemini-generated checklists, and hierarchical aggregation are held fixed; only the final evaluator changes.
Table~\ref{tab:gpt54_results} reports GPT-5.4 scores.
GPT-Image-2, Nano Banana Pro, and Nano Banana 2 remain the top three under both judges, while Seedream 4.0, GPT-Image-1.5, and Seedream 5.0 reorder. Together, these results indicate that the main conclusions are stable under a non-Gemini evaluator and are not explained by same-family coupling. Empirically, the coupling between the VLM evaluator and the image generator appears limited: model-quality differences dominate over generator–judge family membership.
\begin{table}[H]
\centering
\small
\setlength{\tabcolsep}{3.2pt}
\caption{Evaluation results using GPT-5.4 as the judge.}
\label{tab:gpt54_results}
\begin{tabular}{lrrrrrr}
\toprule
Model & Following & Consistency & Composition & Minimality & Quality & Overall \\
\midrule
\rowcolor{gray!15}\multicolumn{7}{c}{\textbf{Closed-source Models}} \\
GPT-Image-2 & 89.99 & 81.90 & 92.04 & 76.59 & 95.67 & 86.37 \\
Nano Banana Pro~\cite{nanobananapro2025} & 86.99 & 76.15 & 86.01 & 72.04 & 88.29 & 81.18 \\
Nano Banana 2~\cite{nanobanana2_2026} & 85.80 & 74.34 & 83.25 & 65.24 & 88.00 & 78.41 \\
Seedream 5.0~\cite{seedream2026seedream5lite} & 83.21 & 66.48 & 80.67 & 63.76 & 85.50 & 74.73 \\
Seedream 4.0~\cite{seedream2025seedream} & 84.38 & 66.99 & 81.46 & 64.10 & 87.35 & 75.54 \\
GPT-Image-1.5~\cite{openai2025gptimage15} & 85.28 & 69.17 & 83.08 & 56.18 & 87.98 & 75.15 \\
\midrule
\rowcolor{gray!15}\multicolumn{7}{c}{\textbf{Open-source Models}} \\
FLUX.2~\cite{blackforestlabs2025flux2} & 45.69 & 42.99 & 52.16 & 46.14 & 62.90 & 48.44 \\
Qwen-Image-Edit~\cite{wu2025qwenimagetechnicalreport} & 47.42 & 41.05 & 53.45 & 42.83 & 61.94 & 47.87 \\
BAGEL~\cite{deng2025bagel} & 39.18 & 35.33 & 37.87 & 47.48 & 39.51 & 39.58 \\
OmniGen2~\cite{wu2025omnigen2} & 33.00 & 19.07 & 34.40 & 19.45 & 41.56 & 28.01 \\
\bottomrule
\end{tabular}
\end{table}

\subsection{Human Evaluation Protocol}
\label{app:human_protocol}
The human evaluation uses 200 benchmark cases and four generators: Nano Banana Pro, Seedream 5.0, FLUX.2, and Qwen-Image-Edit. Each generator contributes one output per case, for 800 generated outputs. The evaluation set contains 5,924 checklist decisions. Five annotators independently judge each evaluated checklist item. They inspect the textual instruction, all reference images, and the generated output, and answer ``Y'' or ``N'' according to whether the checklist item is satisfied, using the same evaluation instructions as the VLM judges. Generator names and identities are hidden, and the order of model outputs is randomized. The final human label is the majority vote of the five annotators. Human inter-annotator agreement is Krippendorff's $\alpha=0.88$.

\begin{table}[H]
\centering
\caption{Task distribution of the 200 human-evaluated benchmark cases.}
\label{tab:human_task_distribution}
\begin{tabular}{lr}
\toprule
Task family & Cases \\
\midrule
Entity Fusion & 70 \\
Attribute Transfer & 60 \\
Multi-Element Extraction & 20 \\
Infographic Manipulation & 30 \\
Visual Reasoning & 20 \\
\midrule
Total & 200 \\
\bottomrule
\end{tabular}
\end{table}

\subsection{Checklist Validity Audit}
\label{app:checklist_audit}
Five PhD-level human auditors examine 2,962 checklist items from 100 benchmark cases covering all five task families. The audit checks whether each item is incorrect, ambiguous, or unanswerable; an item is valid only if it falls into none of these failure categories. Table~\ref{tab:checklist_audit} gives the reported item counts. Of the 2,962 items, 2,925 are valid and 37 are invalid, giving a validity rate of 98.75\% and an error rate of 1.25\%.

\begin{table}[H]
\centering
\caption{Checklist validity audit: reported counts by evaluation dimension.}
\label{tab:checklist_audit}
\begin{tabular}{lrrr}
\toprule
Dimension & Total items & Invalid & Error rate \\
\midrule
Following & 491 & 7 & 1.43\% \\
Consistency & 780 & 11 & 1.41\% \\
Composition & 634 & 6 & 0.95\% \\
Minimality & 584 & 9 & 1.54\% \\
Quality & 473 & 4 & 0.85\% \\
\midrule
Total & 2,962 & 37 & 1.25\% \\
\bottomrule
\end{tabular}
\end{table}

\subsection{Controlled Real-versus-Synthetic Inputs}
\label{app:real_synthetic}
We select 100 cases whose original references are all real images. Each reference is captioned, and Nano Banana Pro synthesizes a corresponding image from the caption. This yields paired real-input and synthetic-input versions with the same task structure and instruction semantics. Nano Banana Pro and Seedream 5.0 are evaluated with the same checklist protocol and Gemini-3-Pro judge.

\begin{table}[H]
\centering
\small
\setlength{\tabcolsep}{3.5pt}
\caption{Paired input-source experiment on 100 cases. Scores are percentages under Gemini-3-Pro.}
\label{tab:real_synthetic}
\begin{tabular}{llrrrrrr}
\toprule
Model & Inputs & Following & Consistency & Composition & Minimality & Quality & Overall \\
\midrule
Nano Banana Pro & Real & 95.37 & 88.39 & 80.19 & 85.64 & 86.67 & 87.01 \\
 & Synthetic & 95.18 & 88.68 & 85.54 & 80.01 & 88.48 & 87.44 \\
Seedream 5.0 & Real & 91.88 & 77.16 & 84.62 & 78.83 & 83.37 & 82.93 \\
 & Synthetic & 90.78 & 76.37 & 87.53 & 75.76 & 88.82 & 83.16 \\
\bottomrule
\end{tabular}
\end{table}

Overall increases by 0.43 points for Nano Banana Pro and 0.23 for Seedream 5.0; the gap changes from 4.08 to 4.28 points. Dimension-level changes are mixed. Both models show small Overall gains, with the performance gap changing by 0.20 points.

\subsection{Repeated Runs on Multi-Element Extraction}
\label{app:run_variance}
We repeat generation and judging independently three times on all 101 Multi-Element Extraction cases. Table~\ref{tab:extraction_runs} reproduces the reported per-run scores, means, and standard deviations. Nano Banana Pro and Nano Banana 2 lead in all three runs; Seedream 4.0 and Seedream 5.0 exchange positions in the third run.

\begin{table}[H]
\centering
\caption{Reported Multi-Element Extraction scores for three independent generation and judge runs, in percent.}
\label{tab:extraction_runs}
\begin{tabular}{lrrrrr}
\toprule
Model & Run 1 & Run 2 & Run 3 & Mean & Std \\
\midrule
Nano Banana Pro & 87.2 & 86.8 & 84.7 & 86.2 & 1.37 \\
Nano Banana 2 & 85.6 & 85.7 & 83.7 & 85.0 & 1.12 \\
Seedream 4.0 & 83.1 & 83.2 & 83.2 & 83.2 & 0.05 \\
Seedream 5.0 & 81.3 & 82.5 & 84.3 & 82.7 & 1.51 \\
\bottomrule
\end{tabular}
\end{table}

\subsection{Limitations}
\label{app:limitations}
Although \name{} covers diverse multi-reference generation settings, constructing realistic and semantically coherent multi-reference cases remains challenging for certain tasks, especially reasoning-intensive and infographic-oriented scenarios. As a result, some challenge settings may still have limited coverage compared with common entity fusion and attribute transfer tasks.

In addition, except for Entity Fusion, Attribute Transfer, and Analogical Reasoning, most subsets in \name{} are constructed using only two reference images. This design choice is intentional: rather than maximizing the number of reference inputs, we focus on evaluating the atomic capabilities of current models, such as fine-grained attribute extraction, symbolic editing, and reasoning-aware generation, under controlled settings.

The benchmark construction pipeline also relies on substantial manual filtering. Current multimodal models still make mistakes when verifying subtle visual differences, distinguishing source-specific details across multiple references, or assessing fine-grained attributes. Manual filtering improves the reliability and authenticity of the benchmark, but also limits the scalability of dataset construction.

Finally, although our checklist-based evaluation improves evaluation consistency and interpretability, automatic evaluation cannot fully replace human judgment. Certain aspects of image quality, such as artistic preference, creativity, and cultural nuance, remain difficult to evaluate using deterministic yes/no questions alone.

\subsection{Compute Resources}
\label{app:compute_resources}

\name{} primarily evaluates model inference rather than training new generative models. The local computational cost therefore comes mainly from generating outputs for open-source baselines and organizing the benchmark artifacts. For open-source models, inference was conducted on GPUs with 80GB of memory to support high-resolution image generation and editing. The exact memory footprint varies across models because the evaluated systems differ in parameter count, precision, and reference-image handling. For proprietary generation models, we accessed the official APIs provided by the model developers; the underlying hardware, memory allocation, and serving details are not disclosed by the providers.

Checklist generation and checklist-based judging were also performed through API calls to the judge model rather than through locally hosted judge inference. As a result, these two stages do not require additional local GPU resources; their practical cost is dominated by API latency, rate limits, and the number of multimodal requests. The evaluation pipeline is embarrassingly parallel across benchmark samples and models. In practice, wall-clock runtime can be reduced by distributing open-source model inference across multiple GPU workers and distributing API-based stages across available API workers.

The benchmark itself does not require model training or hyperparameter search. Reproducing the main evaluation requires storing the benchmark images, generated model outputs, checklist files, and judge responses. We will release scripts that record model identifiers, prompts, input image order, generation parameters when available, and judge outputs, so that future users can audit both the generated images and the checklist-level decisions.

\subsection{Broader Impacts}
\label{app:broader_impacts}

\name{} can have positive societal impact by improving the transparency and reliability of multimodal image generation systems. Multi-reference image generation is increasingly used in design, education, advertising, entertainment, and personal creative workflows. A benchmark that diagnoses whether models preserve identities, transfer attributes correctly, follow instructions, render text accurately, and compose references coherently can help developers understand failure modes before deploying such systems in user-facing settings. More granular evaluation can also reduce over-reliance on single holistic scores, which may hide important errors.

The benchmark may also benefit the research community by making model capabilities easier to compare. Because \name{} decomposes evaluation into interpretable checklist items, researchers can identify whether a model fails due to reference neglect, poor text rendering, inconsistent style transfer, weak visual reasoning, or low-level artifacts. This can guide safer and more targeted model improvement, and can help users choose appropriate systems for different creative tasks.

At the same time, better evaluation of image generation models can indirectly accelerate progress in systems that may be misused. More capable multi-reference generation could be used to create misleading media, impersonation content, deceptive advertisements, or other synthetic images that are difficult for viewers to verify. Our work does not introduce a new generative model, and it does not provide tools for bypassing safety filters. Nevertheless, evaluation benchmarks can influence model development, so we encourage responsible use of \name{} for research, transparency, and safety-oriented model assessment.

Potential mitigations include responsible dataset release, clear documentation of intended use, exclusion of unsafe or inappropriate content, and encouraging model providers to pair improved generation capabilities with provenance tools, watermarking, content moderation, and misuse monitoring. We also encourage future work to study fairness and cultural coverage in multi-reference generation, since reference-conditioned tasks may behave differently across identities, languages, artistic traditions, and visual domains.

\subsection{New Asset Documentation}
\label{app:new_assets}

\name{} introduces a new benchmark asset consisting of task metadata, input images, natural-language instructions, generated checklist items, and evaluation annotations. Each sample records the task family, subtask setting, image order, image roles, content types, and other metadata. This documentation is intended to make the benchmark auditable and to reduce ambiguity in multi-reference settings, where the ordering and role of input images are critical for reproducible evaluation.

The released benchmark will include data-format documentation, field descriptions, and scripts for running checklist generation and checklist-based evaluation. We will also document known limitations, including the reliance on manual filtering, the scope of task categories, and the fact that proprietary model outputs may depend on API versions and availability. The dataset is intended for academic evaluation and diagnostic analysis of image generation systems, not for identity verification, surveillance, or deployment as an automated moderation tool.

Before release, we apply filtering to remove unsafe, inappropriate, or clearly low-quality samples. The benchmark is constructed to avoid collecting private user data, and the annotation process is conducted internally by the authors rather than through public crowdsourcing. We will provide the benchmark with usage guidelines and license information appropriate for research use.

The code is publicly available at \href{\codeurl}{GitHub}, and the benchmark is publicly available at \href{\dataseturl}{Hugging Face}.

\subsection{Prompt Templates for Checklist Generation and Evaluation}

\name{} follows a two-stage checklist-based evaluation protocol. First, a checklist generation prompt asks a multimodal judge to produce atomic, visually verifiable questions for each benchmark sample. Second, an evaluation prompt asks the judge to answer each checklist item for a model output. This design follows the same motivation as binary checklist evaluation: complex multi-reference requirements are decomposed into independent yes/no judgments, making model failures easier to localize.

\clearpage
\begin{promptboxcap}{Checklist Generation Prompt}{Checklist generation prompt used in \name{}. The prompt asks Gemini-3-Pro to generate global- and atomic-level yes/no questions under five dimensions for each multi-reference generation or editing sample.}
{\footnotesize\ttfamily
\begin{flushleft}
\obeylines\obeyspaces
Task: Unified Checklist Generation for Multi-Reference Image Generation / Editing Benchmark

Objective:
You are an expert evaluator designing checklist-based metrics for a multi-reference image generation benchmark.
Given one benchmark sample, generate fine-grained yes/no checklist items that can later be used to evaluate a generated image.

The benchmark includes both editing-like tasks and non-editing multi-reference generation tasks.
In editing-like tasks, one content image is edited using information from reference images.
In non-editing tasks, all input images are references, and the model must extract required elements, attributes, styles, or semantics from them to generate a new image.

Generate objective, visually verifiable checklist items under four dimensions:
1) following
2) consistency
3) composition
4) minimality 
5) quality

Scoring Direction:
Every checklist item must be phrased so that answer "Yes" means this item is satisfied and receives credit.
Answer "No" means this item is not satisfied and receives no credit.
Do not write negatively phrased items where "Yes" indicates failure.

Input JSON fields:
- task\_id in \{fusion, attribute\_transfer, infographic\_manipulation, multiple\_extraction, reasoning\}
- subtask\_id
- instruction\_zh
- instruction\_en
- clear\_instruction (for reasoning task only)
- num\_images
- images: list with \{order, image\_name, role, content\_type\}
  - role=content: image to preserve/edit
  - role=reference: source image to extract details from
  - role may not be accurate. 
  - if all role=reference: this is non-editing multi-reference generation

Dimension guidance:
- following: instruction execution completeness, no missing required operation
- consistency: elements and attributes must match their intended source images (content or reference); no source confusion or attribute mismatch
- composition: multi-source integration should be coherent; spatial relationships, lighting, occlusion, and interactions must be physically and visually plausible
- minimality: no introduction of elements not specified in the instruction; no unintended changes to irrelevant regions or objects; no over-editing beyond the intended scope
- quality: realism, visual coherence, absence of artifacts, readability, and structural integrity

Construction rules:
1) Return checklist questions only, no evaluation result.
2) Each item must be a yes/no question with one atomic criterion.
3) Ground all items in provided instruction + metadata + attached images.
4) Use instruction\_zh and instruction\_en jointly to resolve ambiguity.
5) For reasoning tasks, please use clear\_instruction, instruction\_zh and instruction\_en together.
6) Distinguish editing vs all-reference generation via role.
7) Reference and composition should be detailed for multi-reference tasks.
8) Avoid subjective questions.
9) Avoid redundancy.

Checklist question types:
For each dimension, generate a single list of questions:
- The first 2 questions MUST be global questions (VERY IMPORTANT)
  - High-level, holistic yes/no questions summarizing the dimension
- The remaining questions MUST be atomic checklist items
  - Fine-grained, objective, visually verifiable yes/no questions

Target number of questions (total per dimension):
- following: 4–7
- consistency: 6–12
- composition: 5–10
- minimality: 4–7
- quality: 3–6
The number of questions may be reduced when the task is simple or low-complexity. 

Output format:
Return only a valid JSON object with exactly these keys:
\{
  "following": ["..."],
  "consistency": ["..."],
  "composition": ["..."],
  "minimality": ["..."],
  "quality": ["..."]
\}
No markdown, no extra keys, no extra text.
\end{flushleft}
}
\end{promptboxcap}

\clearpage
\begin{promptboxcap}{Checklist Evaluation Prompt}{Checklist evaluation prompt used in \name{}. The prompt asks the judge model to answer every checklist item with a binary decision, and concise evidence-based reasoning.}
{\footnotesize\ttfamily
\begin{flushleft}
\obeylines\obeyspaces
Task: Unified Checklist-Based Evaluation for Multi-Reference Image Generation / Editing Benchmark

Objective:
You are an expert evaluator for checklist-based assessment.
Your task is to evaluate whether the output image satisfies each checklist item, and return a structured Y/N decision with concise, evidence-based reasoning.

You must evaluate strictly based on:
1) the instruction and metadata,
2) the Input images,
3) the output image,
4) the provided checklist items.

Do not generate new checklist items. Only evaluate the given checklist.

Scoring Rule:
- "Y" means the checklist item is satisfied.
- "N" means the checklist item is not satisfied.
- Each decision must be independent and evidence-based.
- Do not use uncertain labels. If uncertain, choose the more conservative answer based on visible evidence.

Input JSON fields:
- task\_id, instruction\_zh, instruction\_en, num\_images, checklist
- checklist: a dict with exactly five keys:
  - following: list of checklist items
  - consistency: list of checklist items
  - composition: list of checklist items
  - minimality: list of checklist items
  - quality: list of checklist items

Critical constraints:
1) Evaluate checklist items in their original order within each dimension.
2) Output list lengths must exactly match the corresponding checklist list lengths.
3) Reasoning must be concise, specific, and tied to visible evidence.
4) Do not rely on style preference or subjective taste.
5) Do not output extra keys, summaries, markdown, or explanations outside the required JSON object.

Output format:
Return one valid JSON object with exactly three keys:
\{
  "Answer Dict": \{
    "following": ["Y" or "N", ...],
    "consistency": ["Y" or "N", ...],
    "composition": ["Y" or "N", ...],
    "minimality": ["Y" or "N", ...],
    "quality": ["Y" or "N", ...]
  \},
  "Reason Dict": \{
    "following": ["reason for item 1", ...],
    "consistency": ["..."],
    "composition": ["..."],
    "minimality": ["..."],
    "quality": ["..."]
  \}
\}
Return JSON only. No markdown fence, no extra text.
\end{flushleft}
}
\end{promptboxcap}
\clearpage

\clearpage
\section{Qualitative Visualization Results}

\subsection{Visualization Set 1}

\begin{figure}
    \centering
    \includegraphics[
        width=1\linewidth,
        trim=8pt 8pt 8pt 8pt,
        clip
    ]{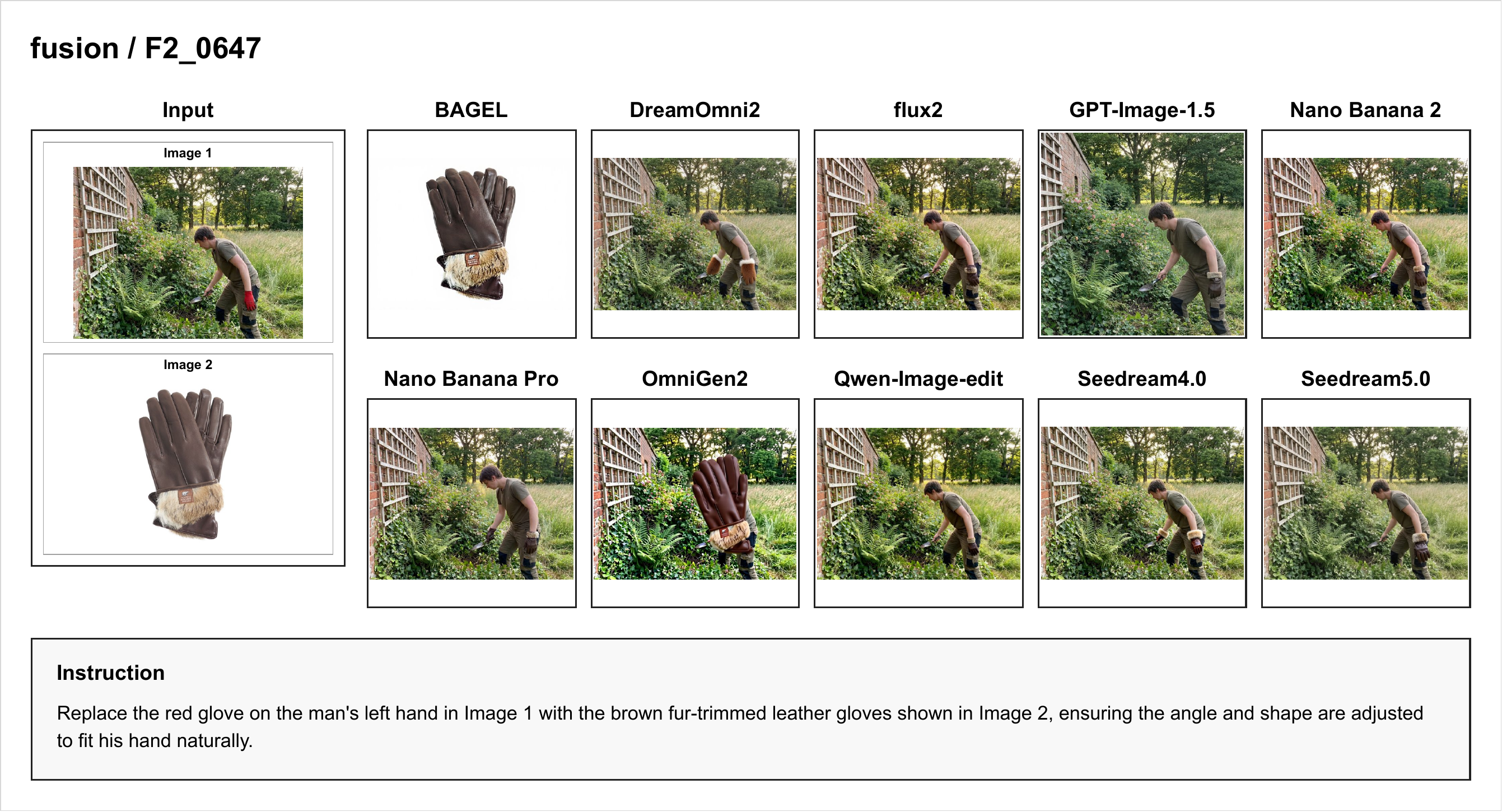}
\end{figure}

\begin{figure}
    \centering
    \includegraphics[
        width=1\linewidth,
        trim=8pt 8pt 8pt 8pt,
        clip
    ]{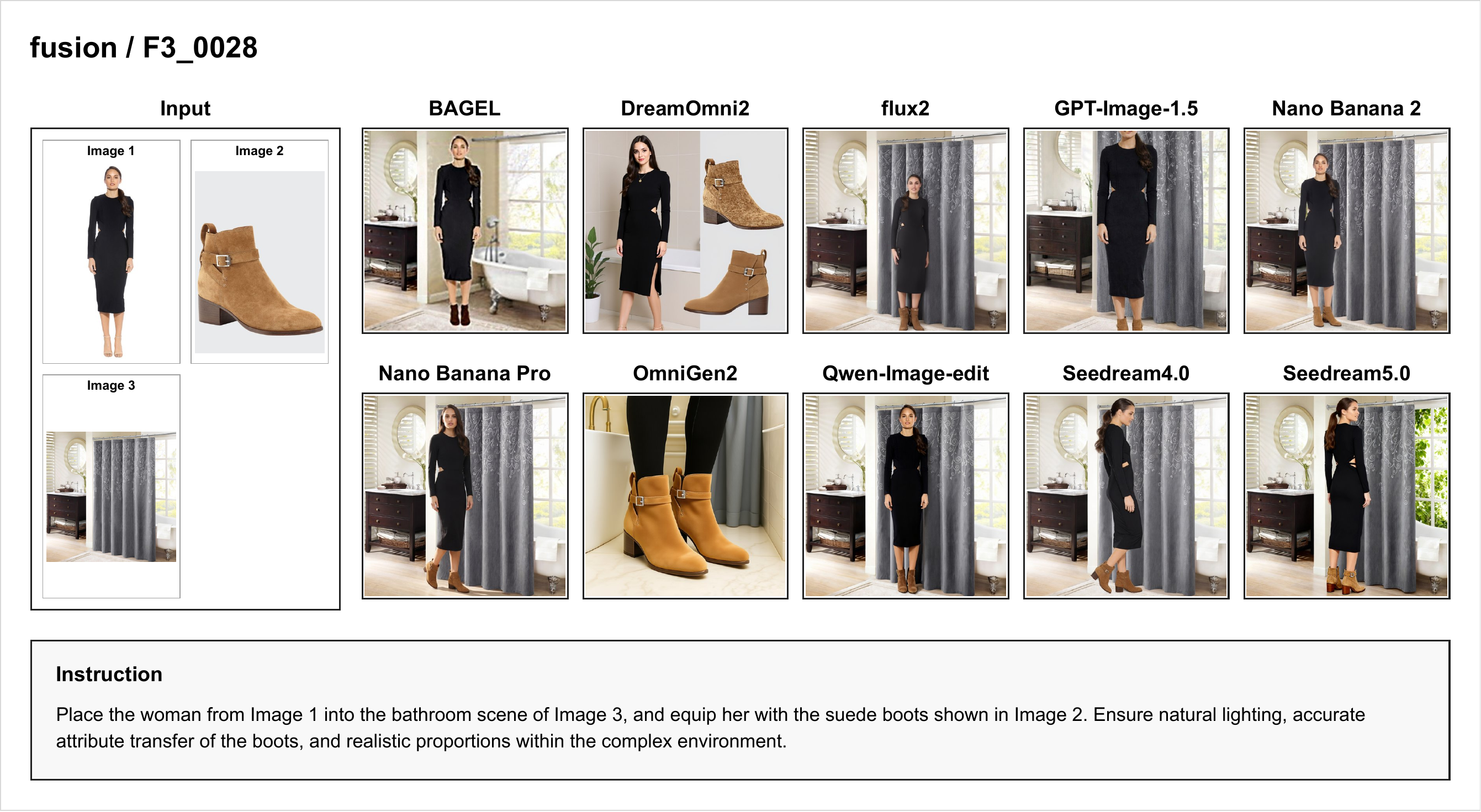}
\end{figure}

\begin{figure}
    \centering
    \includegraphics[
        width=1\linewidth,
        trim=8pt 8pt 8pt 8pt,
        clip
    ]{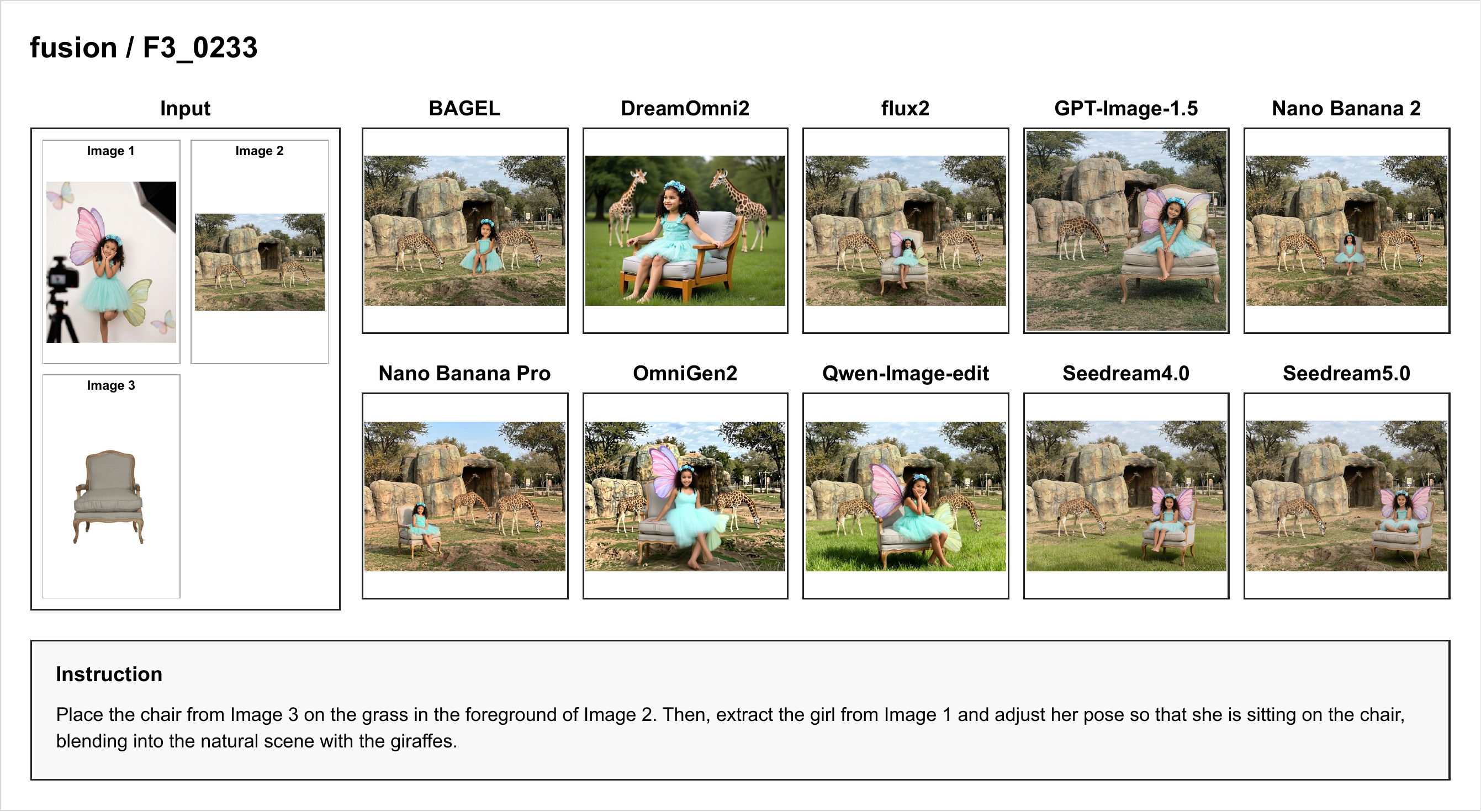}
\end{figure}

\begin{figure}
    \centering
    \includegraphics[
        width=1\linewidth,
        trim=8pt 8pt 8pt 8pt,
        clip
    ]{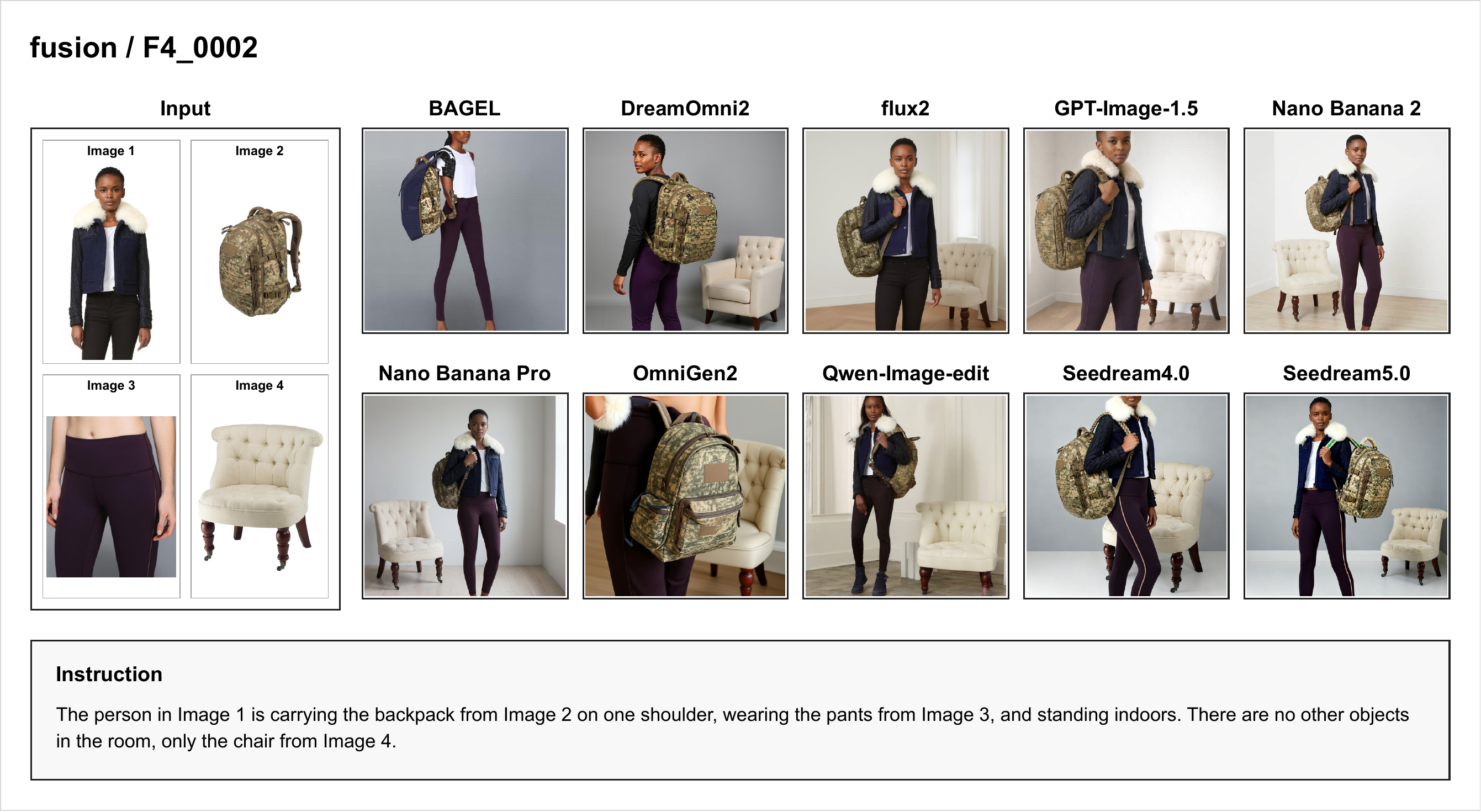}
\end{figure}

\begin{figure}
    \centering
    \includegraphics[
        width=1\linewidth,
        trim=8pt 8pt 8pt 8pt,
        clip
    ]{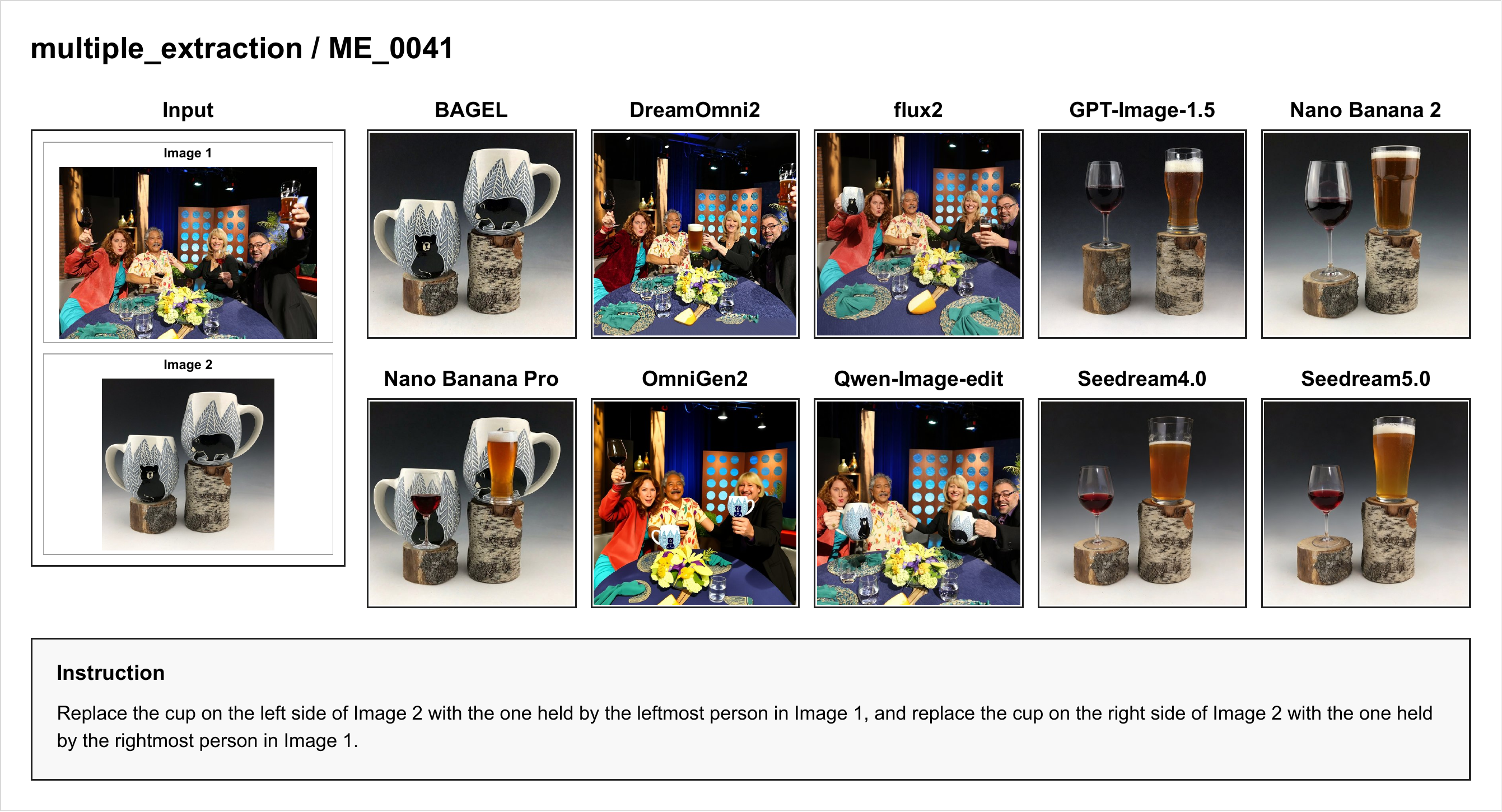}
\end{figure}

\begin{figure}
    \centering
    \includegraphics[
        width=1\linewidth,
        trim=8pt 8pt 8pt 8pt,
        clip
    ]{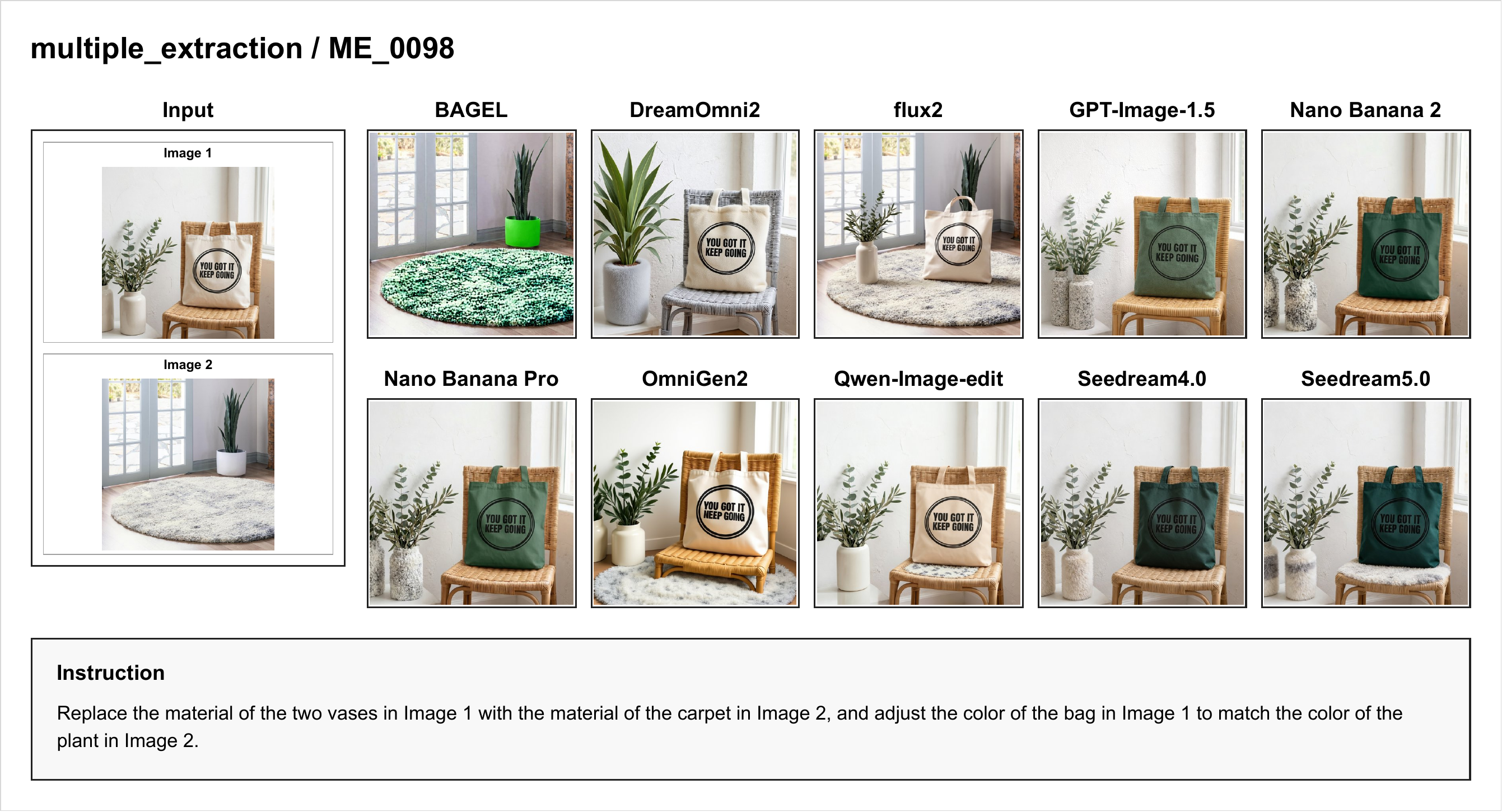}
\end{figure}

\begin{figure}
    \centering
    \includegraphics[
        width=1\linewidth,
        trim=8pt 8pt 8pt 8pt,
        clip
    ]{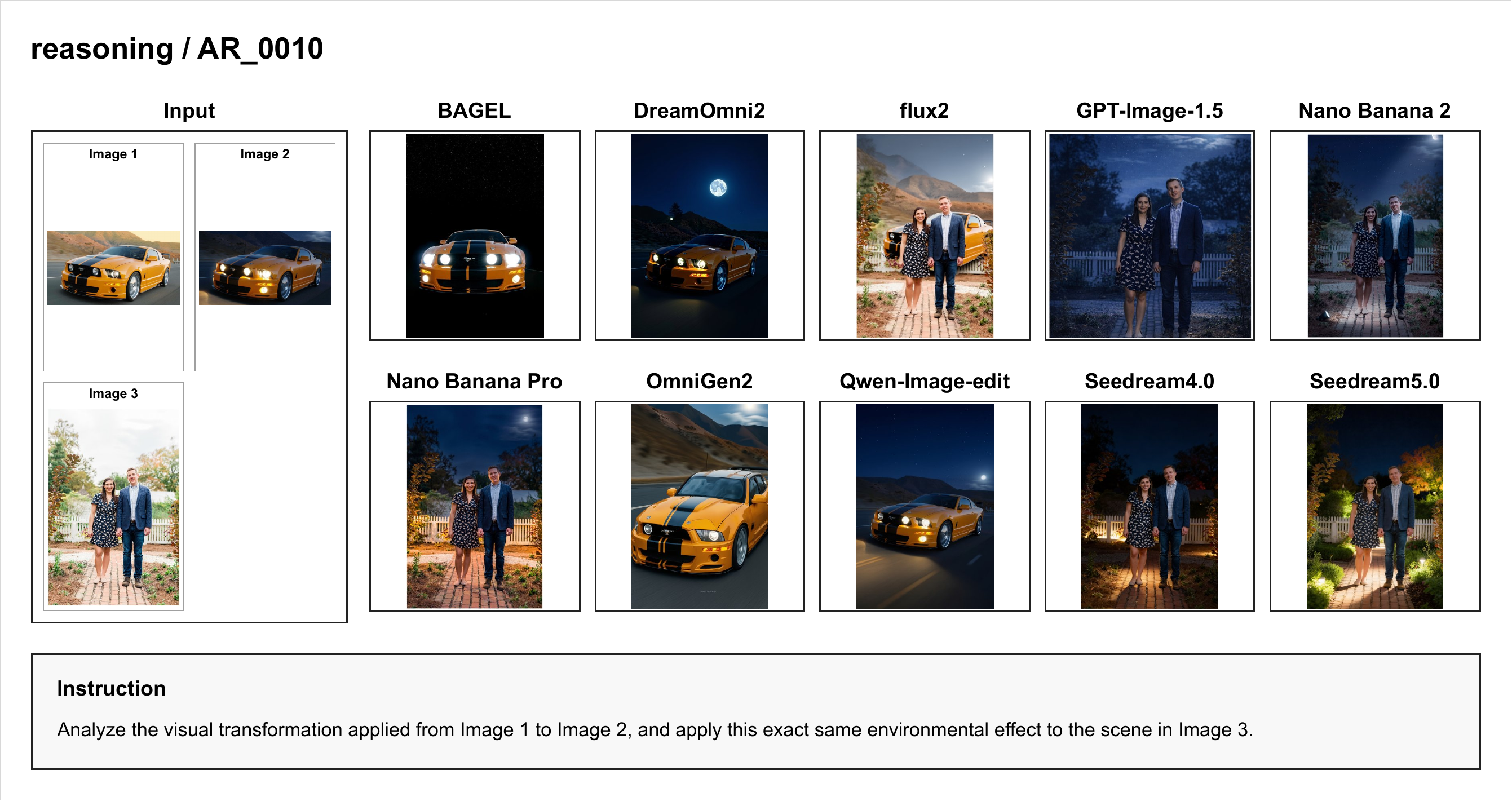}
\end{figure}

\begin{figure}
    \centering
    \includegraphics[
        width=1\linewidth,
        trim=8pt 8pt 8pt 8pt,
        clip
    ]{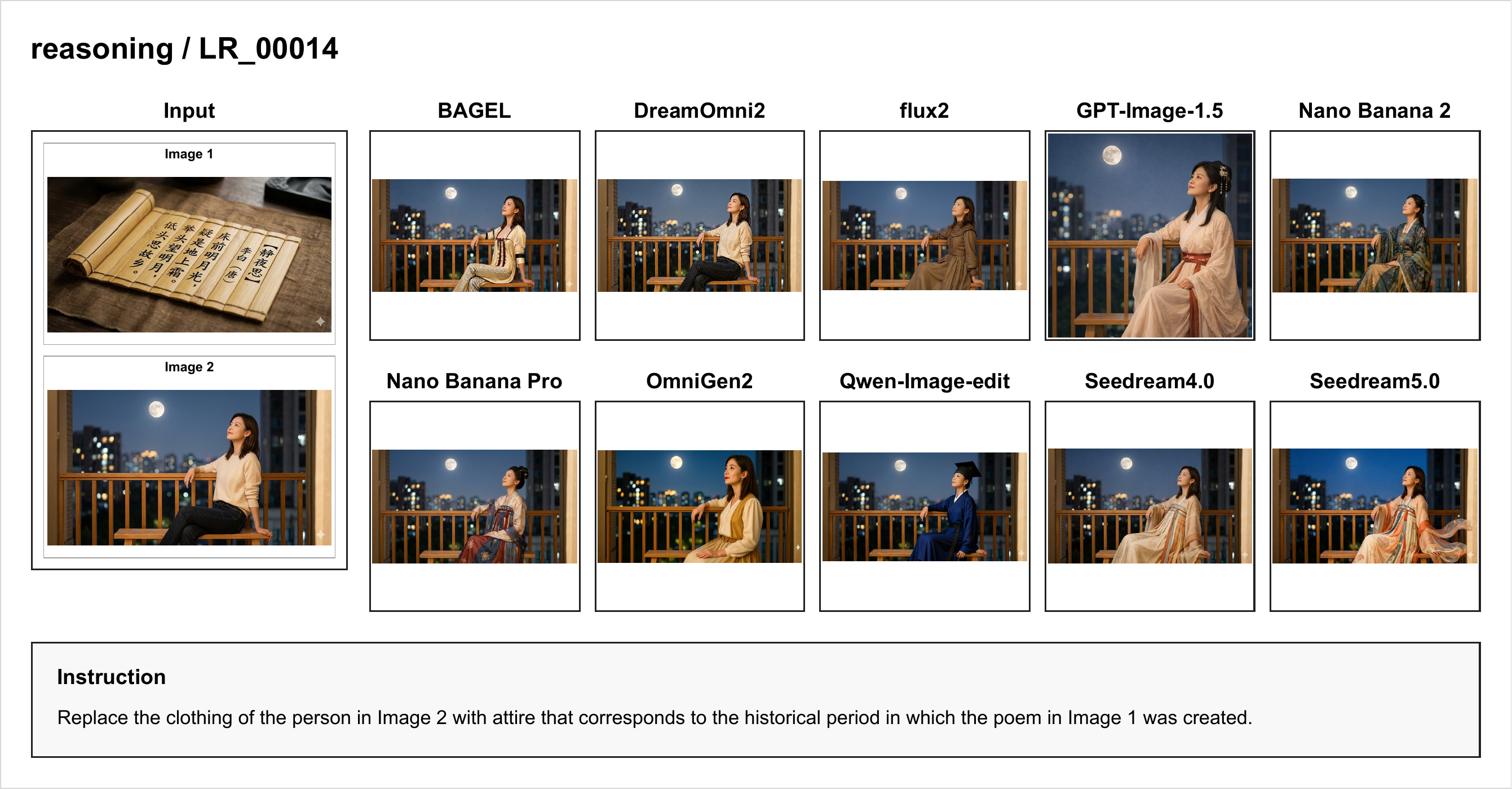}
\end{figure}

\begin{figure}
    \centering
    \includegraphics[
        width=1\linewidth,
        trim=8pt 8pt 8pt 8pt,
        clip
    ]{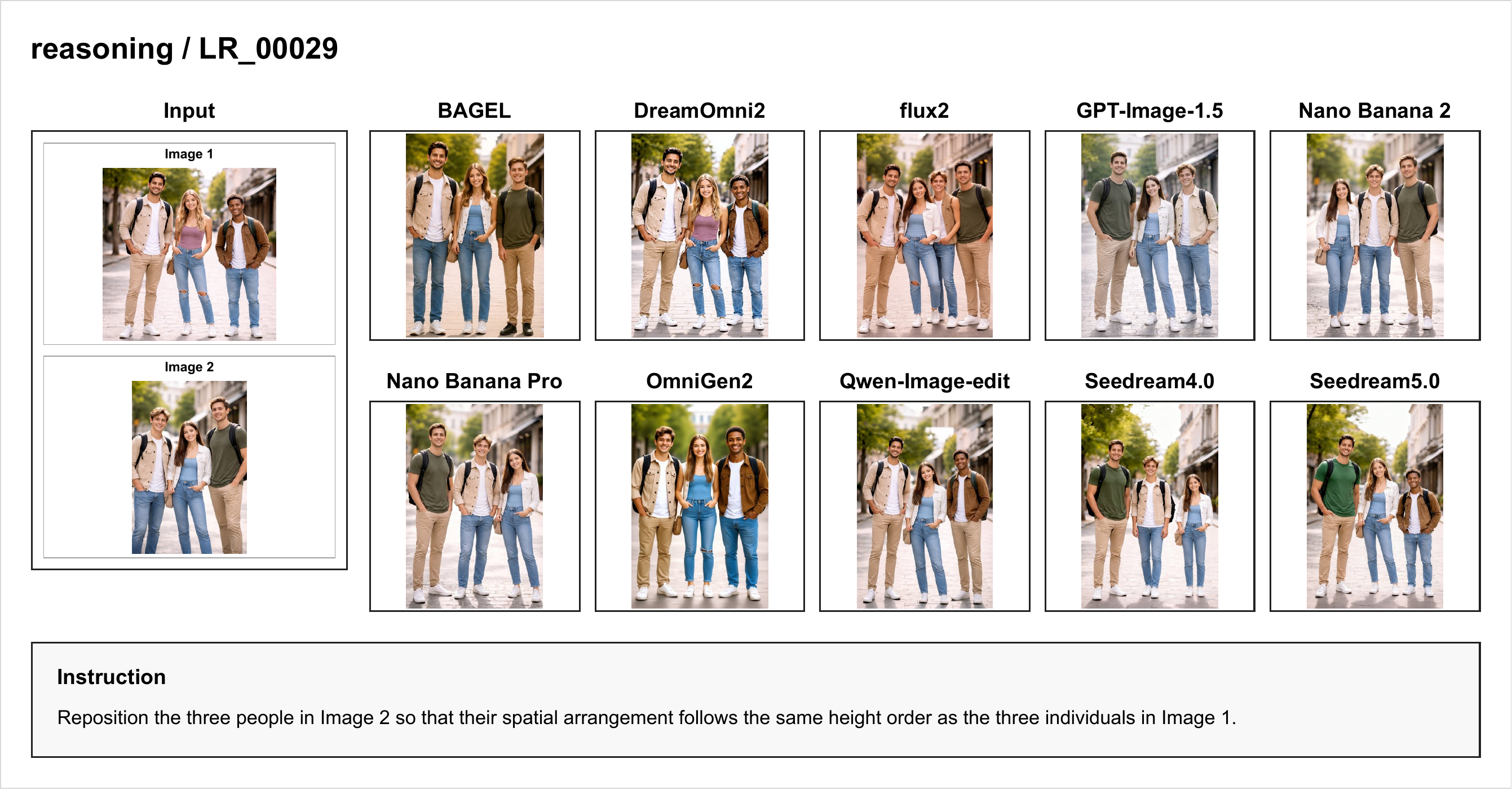}
\end{figure}

\begin{figure}
    \centering
    \includegraphics[
        width=1\linewidth,
        trim=8pt 8pt 8pt 8pt,
        clip
    ]{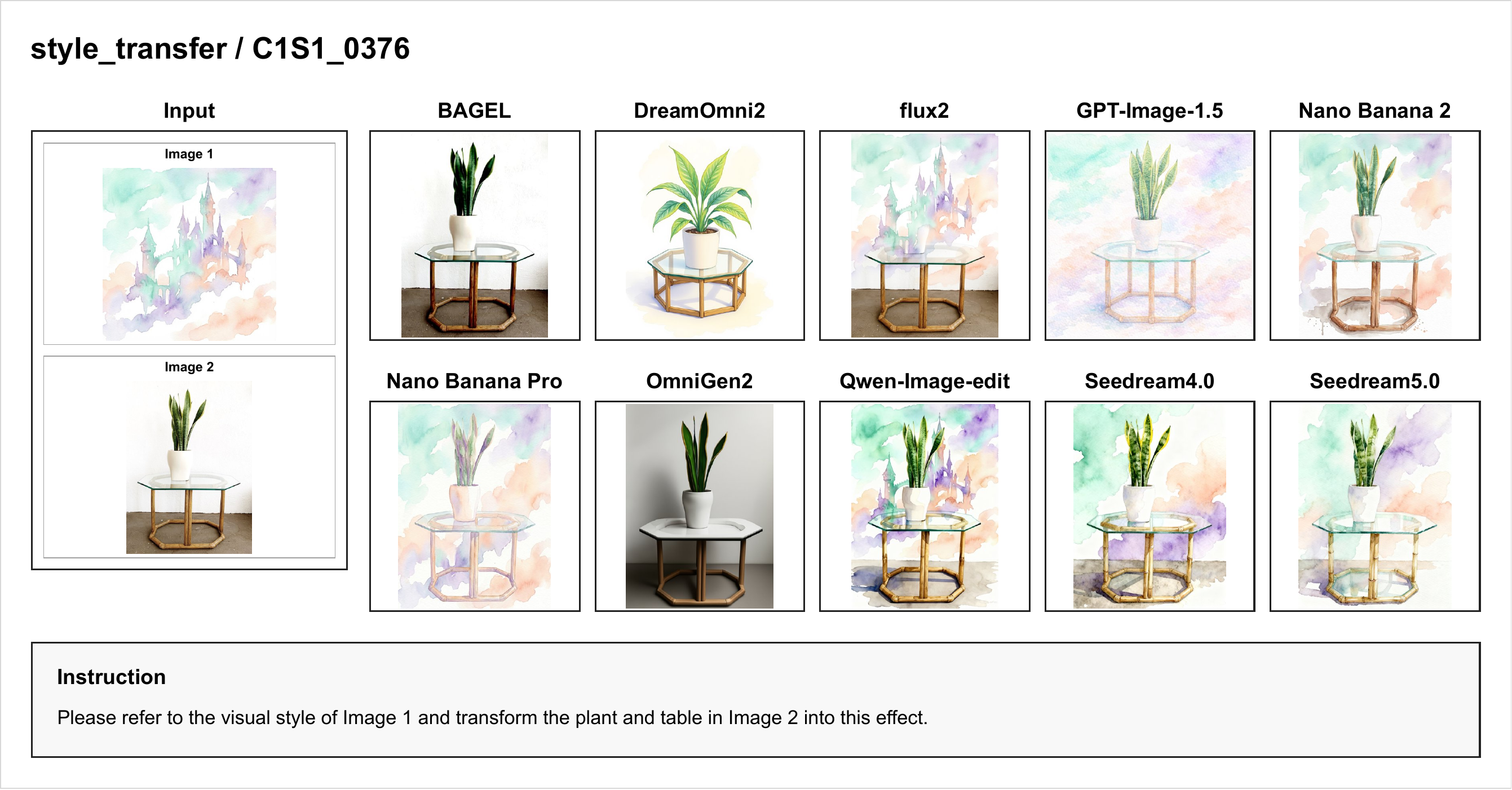}
\end{figure}

\begin{figure}
    \centering
    \includegraphics[
        width=1\linewidth,
        trim=8pt 8pt 8pt 8pt,
        clip
    ]{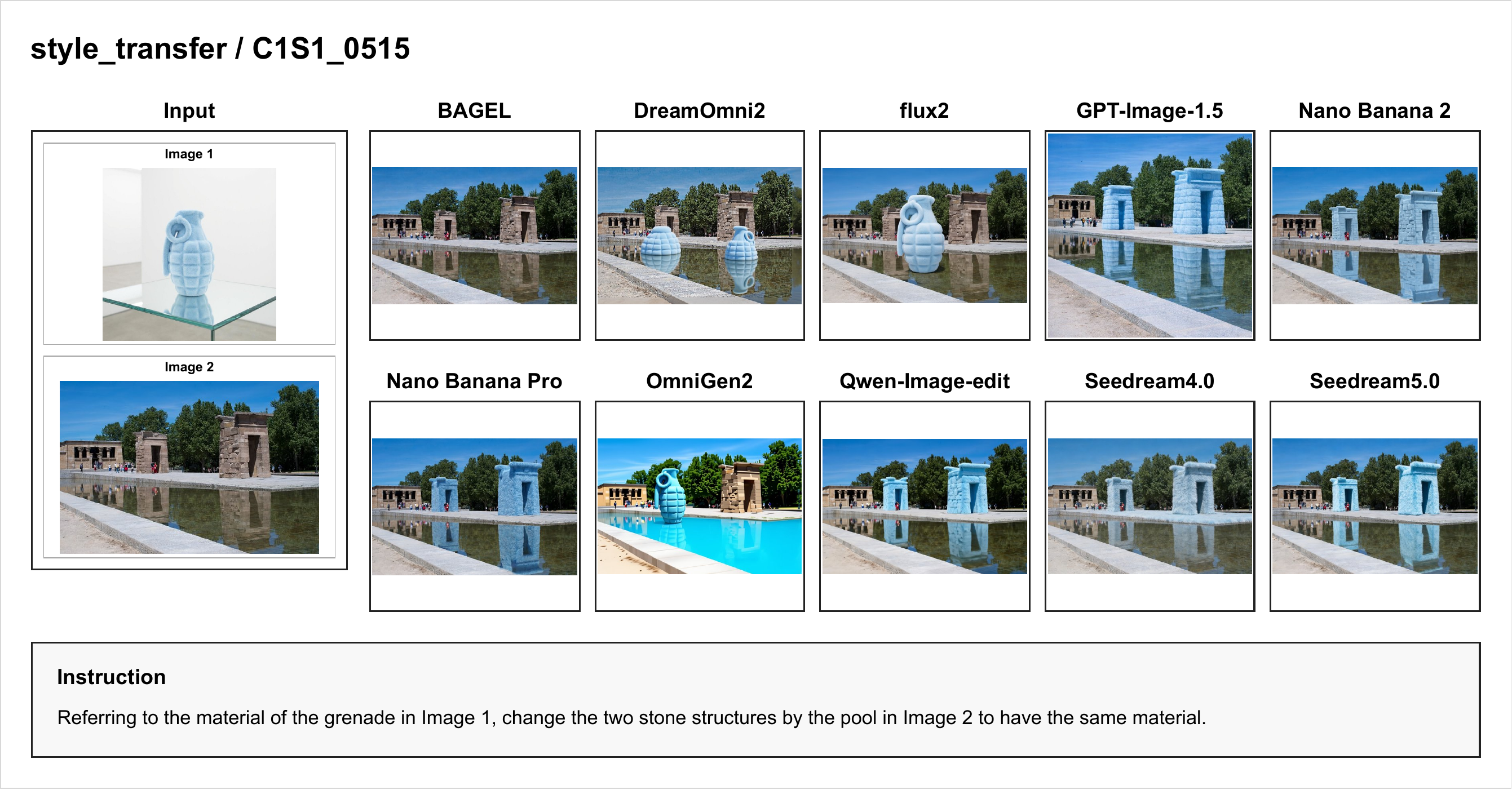}
\end{figure}

\begin{figure}
    \centering
    \includegraphics[
        width=1\linewidth,
        trim=8pt 8pt 8pt 8pt,
        clip
    ]{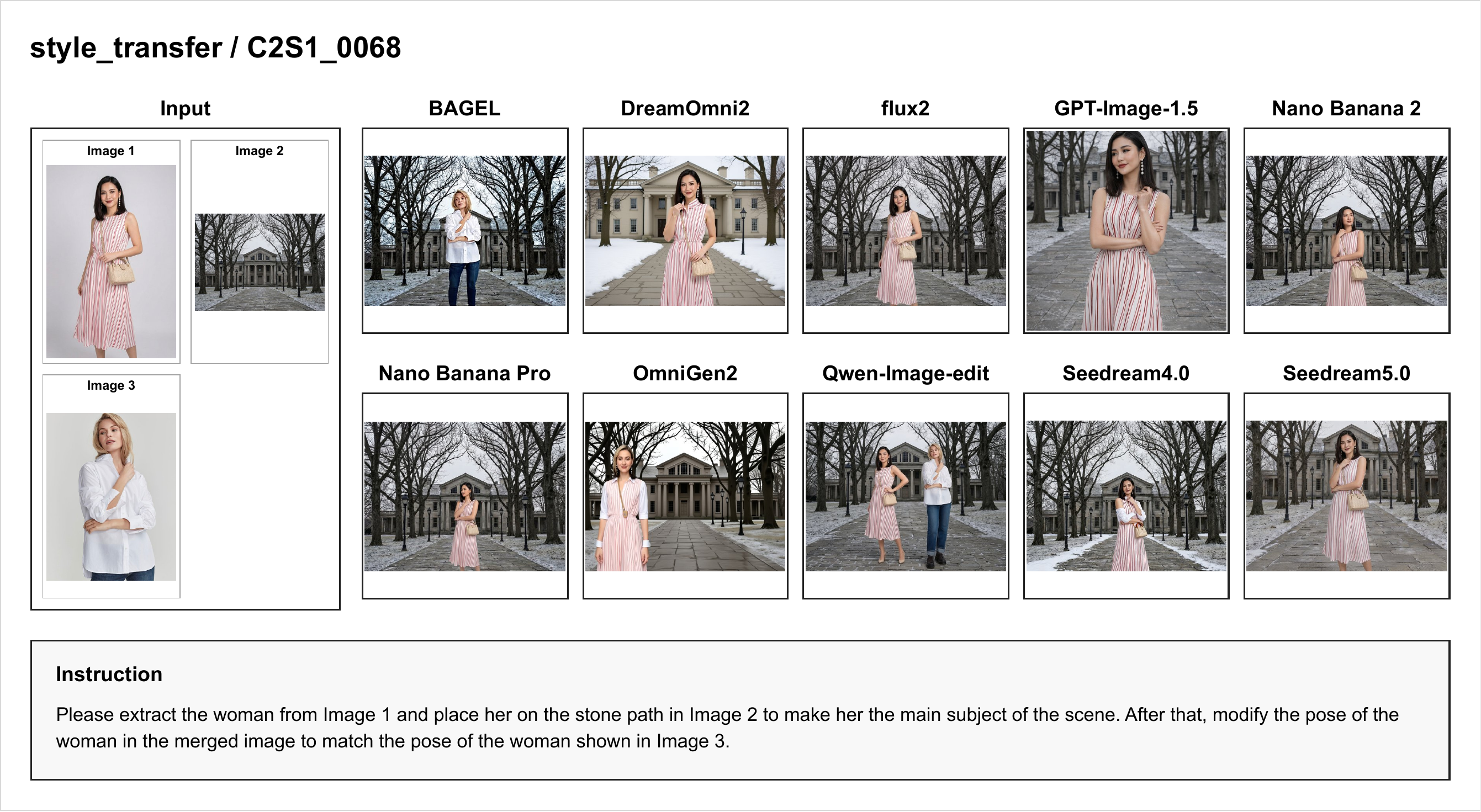}
\end{figure}

\begin{figure}
    \centering
    \includegraphics[
        width=1\linewidth,
        trim=8pt 8pt 8pt 8pt,
        clip
    ]{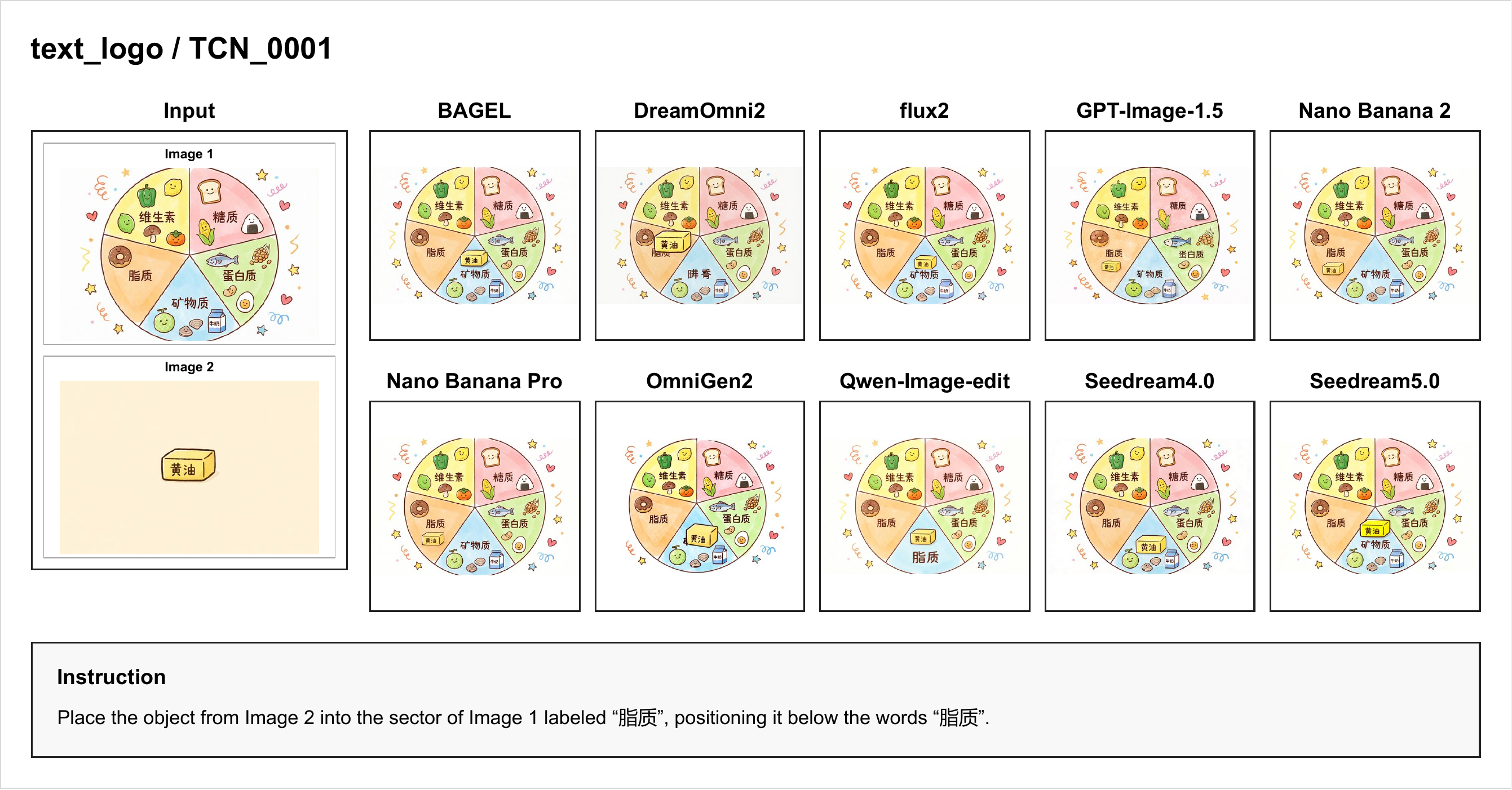}
\end{figure}

\begin{figure}
    \centering
    \includegraphics[
        width=1\linewidth,
        trim=8pt 8pt 8pt 8pt,
        clip
    ]{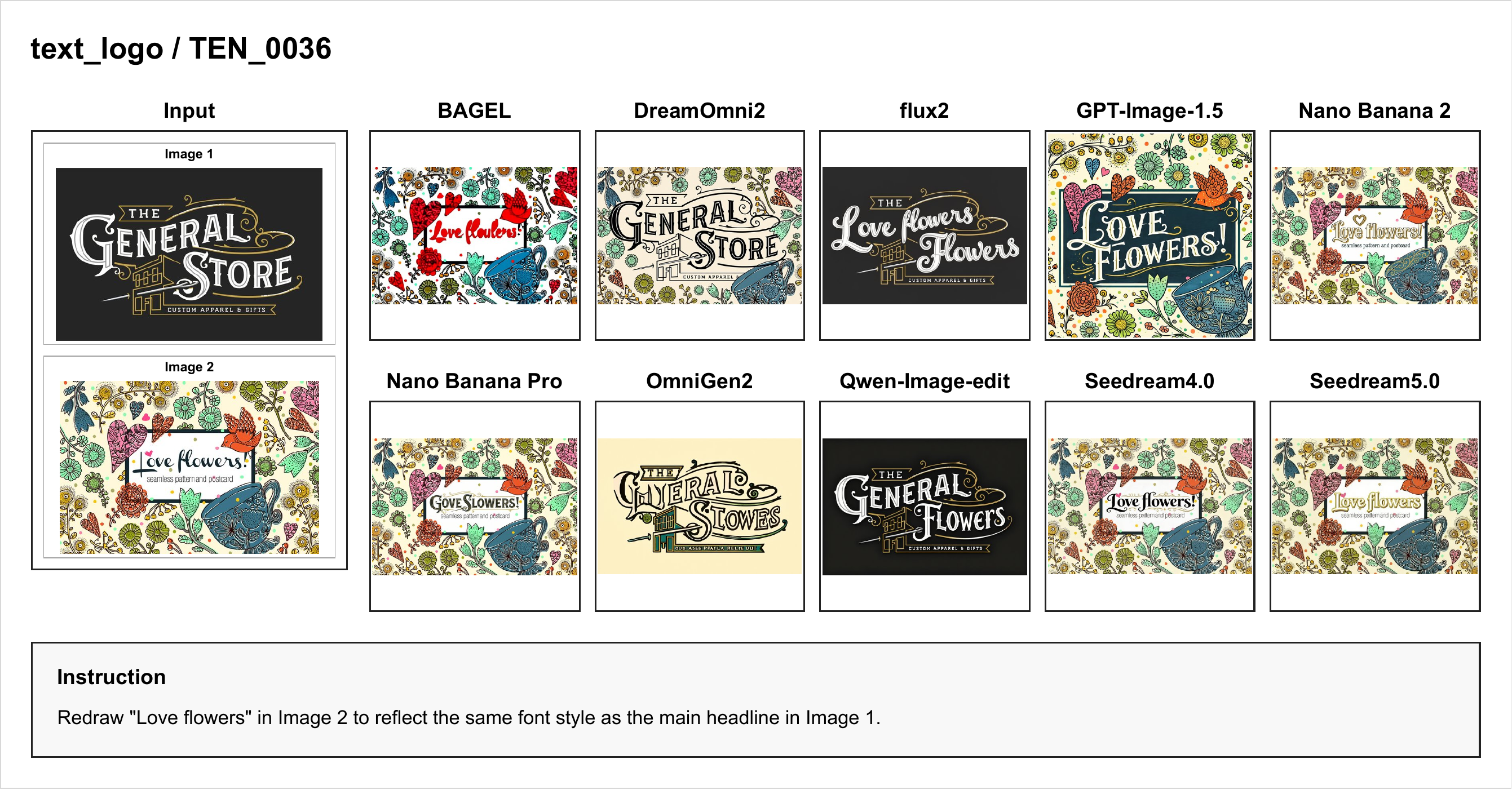}
\end{figure}

\begin{figure}
    \centering
    \includegraphics[
        width=1\linewidth,
        trim=8pt 8pt 8pt 8pt,
        clip
    ]{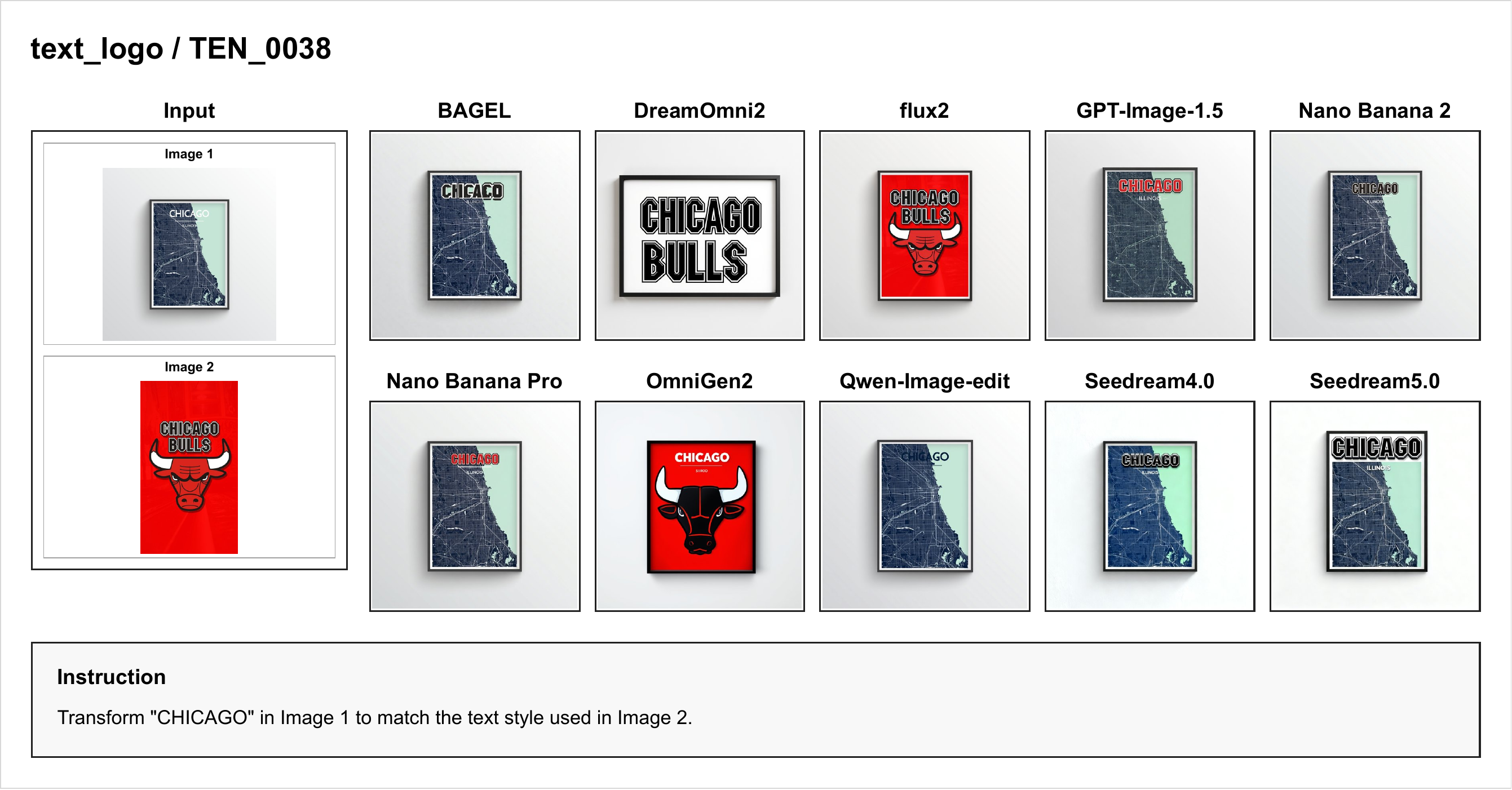}
\end{figure}

\begin{figure}
    \centering
    \includegraphics[
        width=1\linewidth,
        trim=8pt 8pt 8pt 8pt,
        clip
    ]{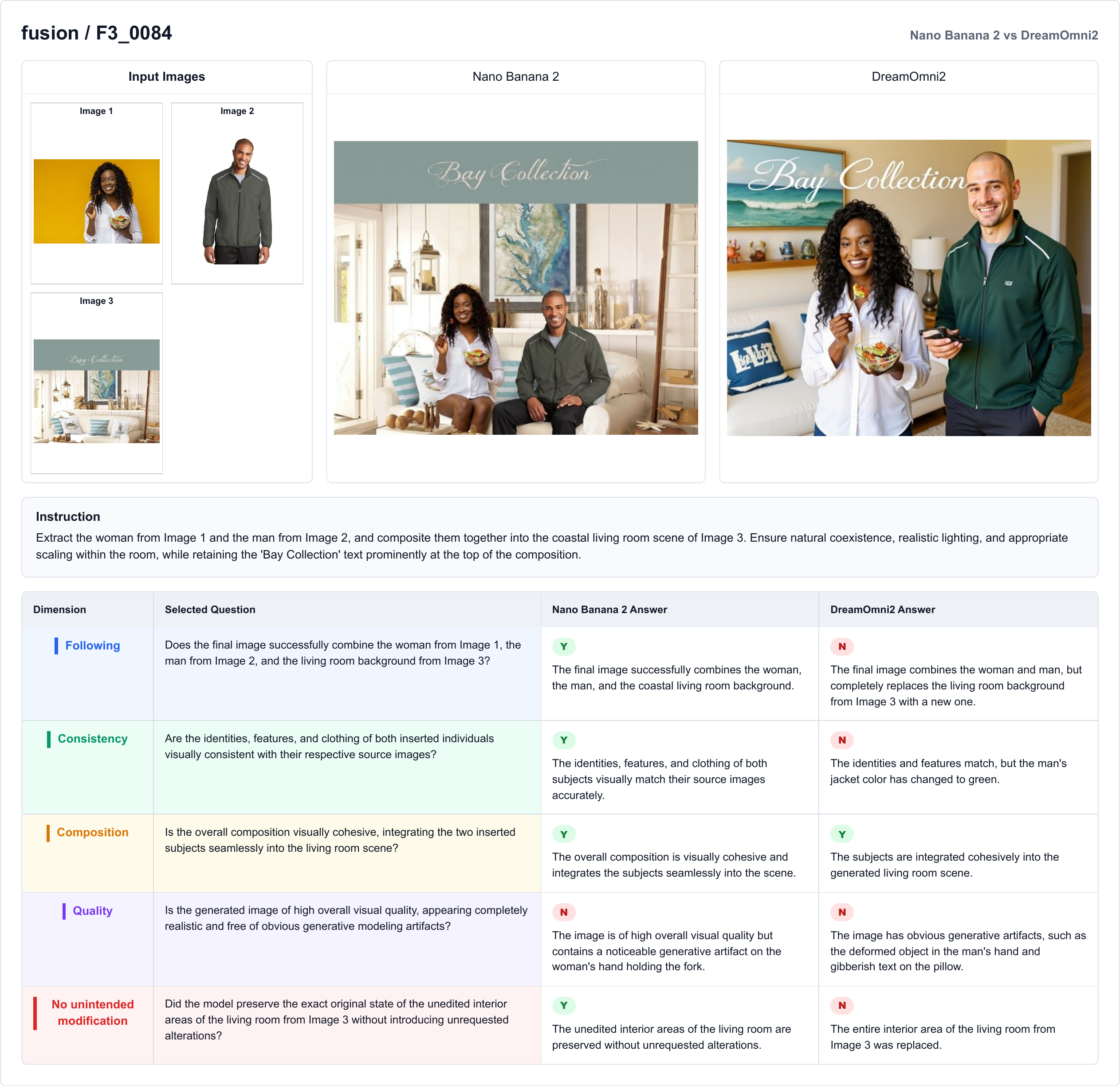}
\end{figure}

\begin{figure}
    \centering
    \includegraphics[
        width=1\linewidth,
        trim=8pt 8pt 8pt 8pt,
        clip
    ]{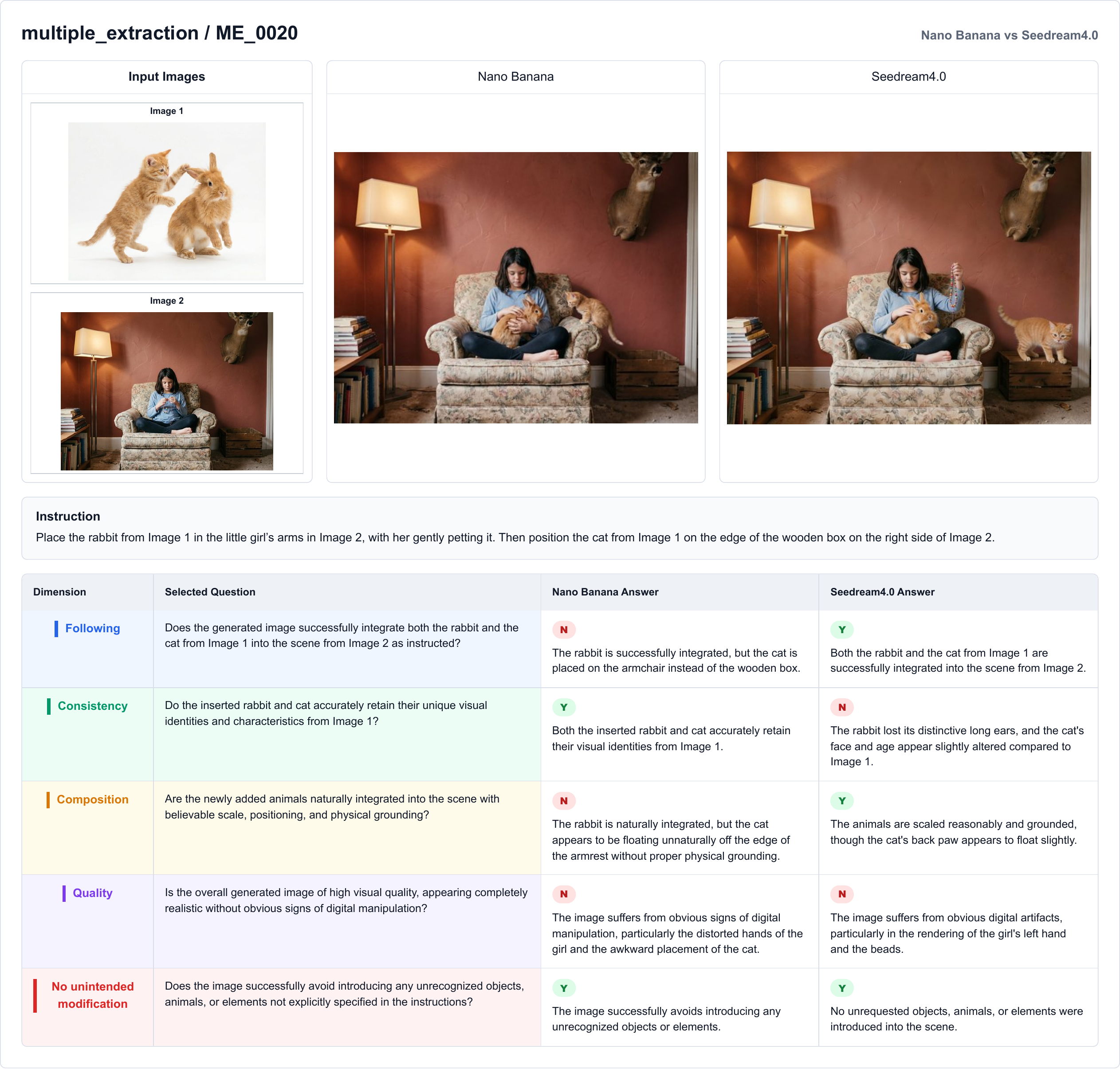}
\end{figure}

\begin{figure}
    \centering
    \includegraphics[
        width=1\linewidth,
        trim=8pt 8pt 8pt 8pt,
        clip
    ]{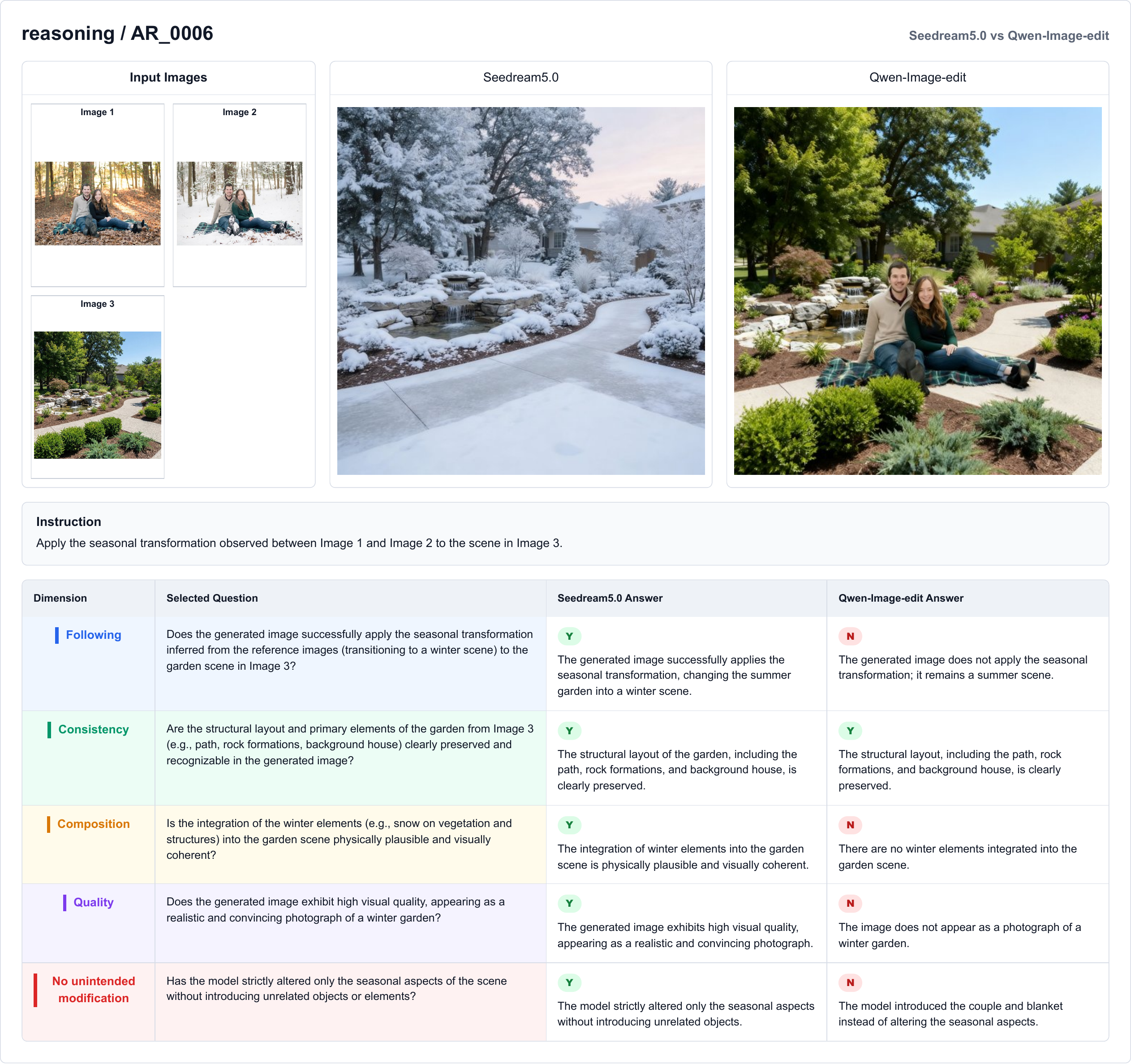}
\end{figure}

\begin{figure}
    \centering
    \includegraphics[
        width=1\linewidth,
        trim=8pt 8pt 8pt 8pt,
        clip
    ]{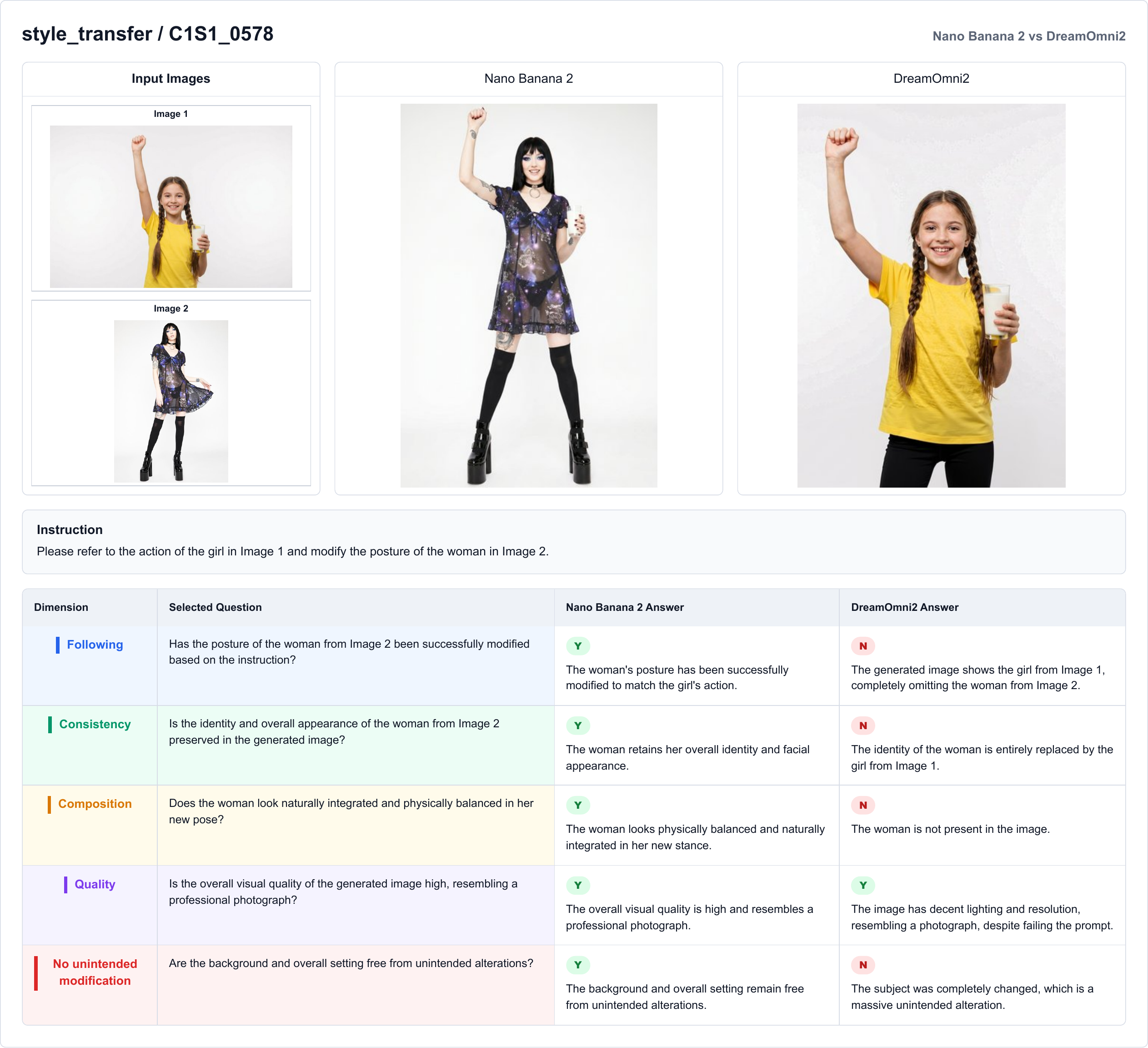}
\end{figure}

\begin{figure}
    \centering
    \includegraphics[
        width=1\linewidth,
        trim=8pt 8pt 8pt 8pt,
        clip
    ]{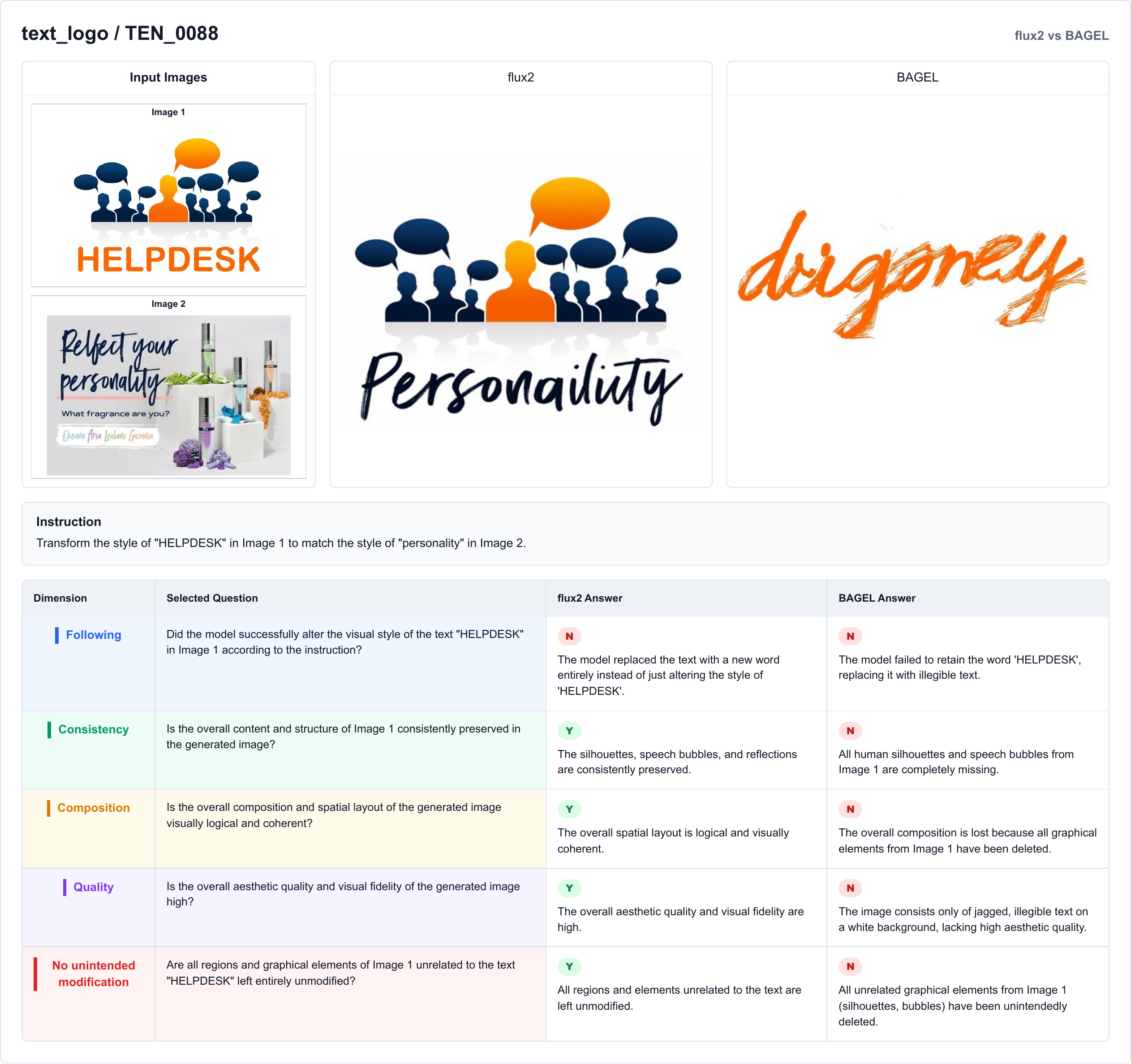}
\end{figure}

\clearpage

\ifdefined\arxivmode\else
  \newpage
  \input{sections/X_checklist.tex}
\fi
\end{document}